\documentclass[sn-mathphys-num]{sn-jnl}
\usepackage{graphicx}%
\usepackage{multirow}%
\usepackage{amsmath,amssymb,amsfonts}%
\usepackage{amsthm}%
\usepackage{mathrsfs}%
\usepackage[title]{appendix}%
\usepackage{xcolor}%
\usepackage{textcomp}%
\usepackage{manyfoot}%
\usepackage{booktabs}%
\usepackage{algorithm}%
\usepackage{algorithmicx}%
\usepackage{algpseudocode}%
\usepackage{listings}%

\usepackage{bookmark}
\usepackage{hyperref}
\hypersetup{pdfauthor={Name}}
\usepackage[T1]{fontenc}
\usepackage{lmodern}
\usepackage{tabularray}
\usepackage{longtable}
\UseTblrLibrary{booktabs}
\usepackage{anyfontsize}
\theoremstyle{thmstyleone}%
\usepackage{mathtools} 
\theoremstyle{thmstyletwo}%

\theoremstyle{thmstylethree}%

\begin{document}

\title[Article Title]{A Survey on the Recent Random Walk-based Methods for Embedding Knowledge Graphs}

\author*[1]{\fnm{Elika} \sur{Bozorgi}}\email{elika.bozorgi@uga.edu}
\author[1]{\fnm{Sakher Khalil} \sur{Alqaiidi}}\email{sakher.a@uga.edu}
\author[1]{\fnm{Afsaneh} \sur{Shams}}\email{afsaneh.shams@uga.edu}
\author[1]{\fnm{Hamid Reza} \sur{Arabnia}}\email{hra@uga.edu}
\author[1]{\fnm{Krzysztof} \sur{Kochut}}\email{kkochut@uga.edu}
\affil{$^{1}$School of Computing, The University of Georgia, Athens, GA, USA}

\abstract{Machine learning, deep learning, and NLP methods on knowledge graphs are present in different fields and 
have important roles in various domains from self-driving cars to friend recommendations on social media platforms. However, to apply these methods to knowledge graphs, the data usually needs to be in an acceptable size and format. In fact, knowledge graphs normally have high dimensions and therefore we need to transform them to a low-dimensional vector space. An embedding is a low-dimensional space into which you can translate high dimensional vectors in a way that intrinsic features of the input data are preserved. In this review, we first explain knowledge graphs and their embedding and then review some of the random walk-based embedding methods that have been developed recently.}


\maketitle

\section{Introduction}
A knowledge graph represents a graph of real-world entities and demonstrates the relationships between them. Knowledge graphs store large amounts of data by connecting large datasets in a structured and meaningful way which leads to data integration and semantic understanding. For instance, \textit{Google Knowledge Graph} uses semantic search information from various sources to enhance its search engine results and display them in a structured manner \cite{steiner2012adding}. In addition, knowledge graphs support knowledge discovery and data exploration by enabling users to discover new relationships and insights from the stored data. For example, in the healthcare field, healthcare professionals can discover correlations between patient demographics and medical conditions and therefore identify best practices and treatments. Another example is in the E-commerce field, where knowledge graphs are used to maintain information about users' purchase histories and different products and therefore offer personalized recommendations \cite{li2020alimekg}.
\newline
\newline
Furthermore, knowledge graphs serve as a foundation for AI and machine learning applications by providing structured data for different purposes such as training models and improving the interpretability of the AI systems. As an example, in the finance area, knowledge graphs are used by integrating different data sources and capturing relationships between entities, and using machine learning and AI techniques to discover patterns and anomalies indicative of fraud \cite{mao2022financial}. 
In another case, knowledge graphs play a crucial role in drug discovery. For example, they integrate large amounts of biomedical data and capture complex relationships between biological entities, facilitate data-driven decision-making, and ultimately accelerate the drug development process \cite{soleymani2023dark}. 
There are many other areas knowledge graphs are used in, such as networking and telecommunications \cite{krinkin2020architecture}, manufacturing \cite{buchgeher2021knowledge}, autonomous vehicles \cite{tezerjani2024real}, smart cities \cite{ahmed2022knowledge} and urban planning \cite{liu2023urbankg}, etc.
\newline
\newline
To take advantage of knowledge graphs in various fields, we would normally use machine learning, deep learning, and AI models on the graph datasets. However, large knowledge graphs have a high dimension, which makes it hard for many machine learning models to work with them. To overcome this problem, we use embedding. Embedding is a representation learning method to map out data to a lower-dimensional vector space, while preserving the main features of the input data \cite{luo2003spectral}. Embedding techniques offer a powerful way to efficiently represent data, leading to semantic understanding, improved model performance, feature extraction, and transfer learning capabilities in machine learning and AI. 
\newline
\newline
Based on \cite{cai2018comprehensive}, there are five major categories for embedding knowledge graphs. These include matrix factorization, generative models, deep learning, graph kernels, and edge reconstruction-based optimization models. Each category includes different techniques for embedding knowledge graphs and several recent ones are summarized in Table 2 for each category. For example, the deep learning category includes famous methods among which are random walk-based ones. Random walks have been used in many models for embedding, due to their important features like exploring graph structures such as local and global context, finding semantically similar nodes, feature extractions, etc.  
In this work, we focus on several important and well-known random walk-based embedding techniques from the deep learning category that have been developed during the recent years.  

\section{Preliminaries}\label{sec2}

In this section, we explain some preliminary concepts used in this review.
\newline
\newline
\textbf{Knowledge graph \cite{sipser1996introduction}}: A knowledge graph is a directed graph whose nodes are entities and edges are relations between entities. It is denoted as $G = (V,E)$ in which $v_i \in V$ are nodes or entities and $e_i \in E$ are edges or relations. Nodes have a type mapping function of $\phi: V \rightarrow T$ where $T$ is the node type, and edges have a type mapping function of $\phi: E \rightarrow R$ where $R$ denotes the edge type.

\hfill\break 
\textbf{Random walk \cite{xia2019random}}: A random walk describes a path consisting of random steps on a mathematical space. It can be denoted as ${\xi_t, t = 0, 1, 2,...}$
where $\xi_t$ describes the position of a random walk after $t$ steps. Since we are using this construct on graphs, here is a brief definition of random walks on graphs in simple words:
a random walk on a graph is a process that begins at a random vertex, and at each step the walk randomly moves to another vertex \cite{spielman2006random}. 

\hfill\break 
\textbf{Skip-gram \cite{tang2015line}}: Skip-gram is an unsupervised algorithm used to find the most related word for a given word. It is a simple neural network with one hidden layer and no activation function. The hidden layer does the dot product between the input vector and the weight matrix. The result of this product is passed to the output vector. Next, a softmax function is applied to the output vector showing the probability of the words appearing in the context. Figure 1 \cite{wu2020toward} describes its architecture.

\hfill\break
 \begin{figure}
   \vspace*{1mm}
   \centering
   \includegraphics[scale=0.5]{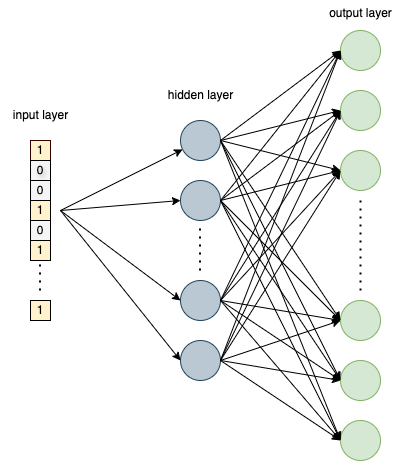}
   \label{fig:figure 1}
   \caption{Skip-gram architecture.}
 \end{figure}
\textbf{Homogeneous Network \cite{cai2018comprehensive}}: Homogeneous Network is a graph denoted as $G = (V, E)$ where $|T^v| = |T^e| = 1$; meaning that all the nodes in $G$ belong to a single type and all the edges in $G$ has a single type, as well.

\hfill\break
\textbf{Heterogeneous Network \cite{li2018link}}: is a graph denoted as $G=(V, E,T)$, where each $v \in V$ and $e \in E$ has a mapping function $\phi(V) = V \rightarrow T_v$ and $\phi(E) = E \rightarrow T_e$ and $T_v$ and $T_e$ denote sets of node and relation types respectively where $|T_v|+|T_e| > 2$.

\section{Methods}\label{sec3}

In this section, we summarize some of the recent algorithms for embedding knowledge graphs which were developed recently. Most of these methods are based on deep learning. Table 1 summarizes these algorithms. 
\hfill\break

\begin{table}[h!]
\centering
\caption{Deep learning algorithms for embedding graphs based on their network type and used method \cite{wang2022survey}.}
\begin{tabular}{|c|c|c|c| }
\hline
Algorithm & Year & Network Type & Random Walk Method \\ [0.5ex] 
\hline  
DeepWalk & 2014 & \multirow{3}{*}{homogeneous}&truncated random walks\\ 
LINE & 2015 & ~ & heterogeneous edges\\
Node2vec & 2016 &~& BFS + DFS based random walks\\ [0.5ex] \hline
PTE & 2017 & \multirow{5}{*}{heterogeneous} & heterogeneous edges \\
Metapath2vec & 2017 & ~ & meta-path based random walks \\ 
Metapath2vec++ & 2019 & ~ & meta-path based random walks \\
Regpattern2vec & 2021 & ~ & regular expression-based random walks \\
Subgraph2vec & 2024 & ~ & truncated random walks \\
 [0.5ex] 
\hline
\end{tabular}

\end{table}

\subsection{Deepwalk \texorpdfstring{\cite{perozzi2014deepwalk}}{}}

DeepWalk is a method for embedding nodes in a network, such as social or biological networks. Prior embedding methods often suffer from scalability issues and fail to capture the structural properties of large-scale networks effectively \cite{tang2009relational,tang2011leveraging}. Deepwalk overcomes these issues in an unsupervised manner. 
\newline
\newline
It is inspired by techniques used in natural language processing, particularly the skip-gram model from the word2vec model \cite{church2017word2vec}. The main idea behind DeepWalk is to treat random walks in a network as "sentences" and learn node embeddings by predicting the context nodes given a target node in these walks.
\newline
\newline
The algorithm generates random walks of fixed length in a network, treating each node as a "word" in a "sentence". It then uses the skip-gram model to learn node embedding by predicting the context nodes for each target node in these random walks. The learned embeddings capture the structural properties of the network, which can be used for various downstream tasks such as node classification, link prediction, and community detection.
\hfill\break

\newpage
\begin{longtblr}[
  caption = {Graph embedding techniques: types, models and algorithms \cite{cai2018comprehensive}},
  label = {tab:test},
]{
   colspec={@{}X[halign=l]X[halign=l]X[halign=l]@{}}
} 
\toprule
Graph Embedding Technique & Model Type & Algorithm\\ 
\midrule
 \SetCell[r=10]{} Matrix factorization & \SetCell[r=5]{} Graph Laplacian Eigenmaps & Isomap \cite{tenenbaum2000global}\\
 & & ARE \cite{lin2005semantic}\\
 & & LPP \cite{he2003locality}\\
 & &SLE \cite{gong2014signed}\\
 & & HSL \cite{sun2008hypergraph}\\
 \cmidrule{2-3}
 & \SetCell[r=5]{} Node proximity matrix factorization & LLE \cite{roweis2000nonlinear}\\
 & & GraRep \cite{cao2015grarep}\\
 & & ULGE \cite{nie2017unsupervised}\\
 & & FONPE \cite{pang2017flexible}\\
 & & SPE \cite{shaw2009structure}\\
\cmidrule{1-3}
\SetCell[r=13]{} Deep Learning & \SetCell[r=8]{}With random walks & DeepWalk \cite{perozzi2014deepwalk}\\
& & node2vec \cite{grover2016node2vec}\\
& & LINE \cite{tang2015line}\\
& & PTE \cite{tang2015pte}\\
& & metapath2vec \cite{dong2017metapath2vec}\\
& & metapath2vec++ \cite{dong2017metapath2vec}\\
& & subgraph2vec \cite{bozorgi2024subgraph2vec}\\
& & regpattern2vec \cite{keshavarzi2021regpattern2vec}\\
\cmidrule{2-3}
& \SetCell[r=5]{}Without random walks & SCNN \cite{bruna2013spectral}\\
& & MoNet \cite{monti2017geometric}\\
& & SDNE \cite{wang2016structural}\\
& & GNN \cite{scarselli2008graph}\\
& & DUIF \cite{geng2015learning}\\
\cmidrule{1-3}
\SetCell[r=7]{} Edge reconstruction & \SetCell[r=4]{} Maximize edge reconstruct probability & PALE \cite{man2016predict} \\
& & APP \cite{zhou2017scalable}\\
& & ESR \cite{xiong2017explicit}\\
& & GAKE \cite{feng2016gake}\\
\cmidrule{2-3}
& \SetCell[r=3]{} Minimize distance-based loss & PLE \cite{ren2016label}\\
& & HEBE \cite{gui2016large}\\
& & IONE \cite{liu2016aligning}\\
& & GraphEmbed \cite{zhang2017regions}\\
\cmidrule{2-3}
\SetCell[r=8]{}Edge reconstruction & \SetCell[r=8]{}Minimize margin-based ranking loss & TransE \cite{bordes2013translating} \\
& & TransH \cite{wang2014knowledge} \\
& & TransR \cite{lin2015learning}\\
& & TransD \cite{ji2015knowledge}\\
& & NTN \cite{socher2013reasoning}\\
& & DistMult \cite{yang2014embedding}\\
& & ComplEx \cite{trouillon2016complex}\\
& & RotatE \cite{sun2019rotate}\\
\cmidrule{1-3}
\SetCell[r=12]{} Graph kernel & \SetCell[r=4]{}Based on graphlet & Graphlet sampling kernel \cite{prvzulj2007biological}\\
& & Graphlet Decomposition Embedding\cite{tong2006center}\\
& & Graph2vec \cite{narayanan2017graph2vec}\\
& & GraRep \cite{cao2016deep} \\
\cmidrule{2-3}
& \SetCell[r=5]{} Based on subtree patterns & Graphlet Kernel \cite{shervashidze2009efficient} \\
& & TreeGCN \cite{ying2018hierarchical}\\
& & Subgraph Isomorphism Network (GIN) \cite{xu2018powerful} \\
& & Subgraph Neural Network \cite{alsentzer2020subgraph}\\
& & Weisfeiler-Lehman(WL)Subtree Kernel \cite{shervashidze2011weisfeiler}\\
\cmidrule{2-3}
& \SetCell[r=3]{}Based on random walks & Graphgan \cite{wang2018graphgan}\\
& & NetMF \cite{qiu2018network} \\
& & GraphSAGE \cite{hamilton2017inductive} \\
\cmidrule{1-3}
\SetCell[r=6]{} Generative model & \SetCell[r=4]{} Embed graph into latent space & VGAE \cite{kipf2016variational}\\
& & GraphRNN \cite{you2018graphrnn} \\
& & Graph-GAE \cite{kipf2016variational}\\
& & Graph-VIN \cite{she2017interactive}\\
\cmidrule{2-3}
& \SetCell[r=2]{} Incorporate semantics for embedding & DGMG \cite{jin2020graph} \\
& & Sem-GAN \cite{cherian2019sem}\\
\SetCell[r=1]{} Generative model & \SetCell[r=1]{} Incorporate semantics for embedding & SE-VAE \cite{fraccaro2016sequential}\\
\cmidrule{1-3}
\end{longtblr}

\textbf{Method:} The algorithm consists of two major parts: 1) random walk generator and 2) update procedure.
\newline
\newline
\textit{Random walk generator:}
The algorithm has two nested loops, the outer loop which represents the number of times ($\gamma$) the walk should start from each vertex $v_i$ of the graph.
DeepWalk starts by generating a random walk $W_{vi}$ in the network from a random node called root ($v_i$). We set the parameter $t$ to control the walk length to have walks of fixed length; however, the walks can be of any length as long as the length is smaller than $t$. The walks are performed starting from each node $\gamma$ times and are entirely random which means they can revisit their root. 
\newline
\newline
\textit{Update procedure:}
The inner loop iterates through the nodes of the graph and starts the walk from each of the vertices.  
Each random walk captures the local neighborhood information around each node. The skip-gram model - commonly used in word embedding techniques like word2vec - is employed to learn embeddings for the nodes.
The actual input for the original skip-gram model are sentences of the words; therefore, we consider the walks as sentences in which the nodes represent the words.
Given a sequence of the nodes by random walk, the skip-gram model aims to predict the context nodes for each target node in the sequence. The objective is to maximize the likelihood of the observing context nodes given the target node. We use skip-gram to update the representations of the nodes according to the following objective function:

\begin{equation*}
  \underset{\Phi}{minimize} -log P_r({v_{i-w}..v{i+w}}\ v_i|\Phi (v_i))  
\end{equation*}
\newline
\newline 
The objective function uses the skip-gram model for learning node embedding and is defined as the log-likelihood of observing the context nodes for each target node across all the random walks. This objective function is optimized using stochastic gradient descent to learn the parameters of the model and is as follows:

\begin{equation*}
    Pr( \{ v_{i-w}...v_{i+w} \} \setminus v_i | \Phi(v_i)) =   {\prod_{\substack{j=i-w\\ j\not = i}}^{i+w}} Pr(v_j | \Phi(v_i)) 
\end{equation*}

where $\phi(v_j)$ represents the embedding of vertex $v_j$.
\newline
\newline
The embeddings are learned iteratively by maximizing the log-likelihood of observing context nodes for each target node in the random walks. This involves updating the embeddings for each node using gradient descent based on the prediction error between the observed and predicted context nodes.

\subsection{LINE \texorpdfstring{\cite{tang2015line}}{}} 

In this section, we review LINE, a method for embedding large networks. LINE is a suitable method for preserving the local pairwise proximity (local structure) between the vertices and dealing with very large networks (\textit{millions of vertices and billions of edges}) with arbitrary types of edges (\textit{directed, undirected, weighted}).
\newline
\newline
The pairwise proximity between vertices includes first-order proximity and second-order proximity. The first-order proximity between two vertices $(u,v)$ is the weight on the edge connecting these vertices $(w_{uv})$. It illustrates the direct similarity between two vertices. For example, people who are friends on social media probably share similar friends. However, the first-order proximity on its own does not preserve the structure of the network. For example, consider a link is missing between two vertices sharing common neighbors. Although these two vertices are very similar, the first-order proximity in this case is 0. Therefore, another parameter that retains the network structure is needed and that is second-order proximity. The second-order proximity between a pair of vertices $(u,v)$ implies the similarity between their neighbors. This helps to identify objects that might not be directly connected but are related through shared neighbors.
\newline
\newline
Combining both of the above proximities, we form LINE; a method for embedding very large networks with arbitrary (directed, undirected, or weighted) edges. 
\hfill\break

\textbf{Method:} First, we combine LINE with each of the mentioned proximities individually and then we combine them. Here is a brief description of the model:
\hfill\break

1. \textbf{LINE with first-order proximity}: 
\newline
\newline
Since joint probability also implies dependencies and relationships between vertices in a graph, we model the first-order proximity for the undirected edge $(v_i, v_j)$ between vertices $v_i$ and $v_j$ which is as follows:

\begin{equation}
    p_1(v_i,v_j) = \frac{1}{1+exp(-\overrightarrow{u_i}^T\cdot \overrightarrow{u_j})}
\end{equation}

where $u_i$ and $u_j$ represent the vector representation of vertices $v_i$ and $v_j$ in a low dimensional space respectively. 
In particular, the joint probability of an edge represents the probability that the two specific nodes are connected by an edge simultaneously. It quantifies the likelihood of a specific edge existing in the graph.
On the other hand, there is another parameter to calculate the likelihood of edges called \textit{empirical probability}. The empirical probability of an edge in a graph is based on the observed data and represents the relative frequency with which a specific edge occurs in the observed graph and is calculated as $\hat {p_1}(i,j) = \frac{w_{ij}}{W}$
where $W = \sum_{(i,j) \in E} w_{ij}$. To preserve the network's first-order proximity, we try to minimize the distance between these two functions:

\begin{equation}
    O_1 = d( \hat p_1 (\cdot , \cdot), p_1(\cdot , \cdot))
\end{equation}
\newline

where $d(\cdot,\cdot)$ is the distance between two vertices.
We choose to minimize the KL-divergence of the two probability distributions which we replace with $d(\cdot,\cdot)$ in the above equation and remove some constants:

\begin{equation}
    O_1 = - \sum_{(i,j) \in E}w_{ij}log p_1(v_i, v_j)
\end{equation}
\newline
\newline
We can represent every node in the d-dimensional space by finding the ${ \{\overrightarrow u_i \}}_{i = 1,..|V|}$ in any undirected graph.
\hfill\break

2. \textbf{Line with $2^{nd}$ order proximity}:
\newline
\newline
The second-order proximity is applicable on both directed and undirected graphs.
This proximity assumes that vertices sharing many other connections are similar to each other. In this proximity, each vertex has two roles: 1. "Vertex" itself and 2. "Context" of other vertices. In the second case, each vertex is considered as a specific context in which the vertices sharing similar distribution over the contexts are considered similar. Therefore, we will have two different representations for vertex $v_i$: $u_i$ and $u_i'$ representing the embeddings of the vertex and context of $v_i$, respectively.
\begin{equation}
    p_2(v_j|v_i) = \frac{1+exp(-\overrightarrow{u'_i}^T\cdot \overrightarrow{u_j})}{\sum_{k=1}^{|V|exp(\overrightarrow{u'_k}^T\cdot \overrightarrow{u_i})}} 
\end{equation}
\newline

where $p (\cdot,v_i)$ is the conditional distribution over the contexts and $|V|$ is the number of the vertices or contexts. In addition, this equation defines a conditional distribution $p_2(\cdot | v_i)$ over the entire set of vertices. To preserve the second-order proximity, we should minimize the distance between $p_2(\cdot | v_i)$ and the empirical distribution $\hat p_2 (\cdot|v_i)$. Therefore:

\begin{equation}
    O_2 = \sum_{i \in  V}\lambda_id(\hat p_2 (\cdot | v_i), p_2(\cdot | v_i))
\end{equation}
\newline

where $d(\cdot,\cdot)$ is the distance between two distributions and $\lambda_i$ denotes the importance of vertex $i$, which can be measured by the degree or estimated through algorithms such as PageRank\cite{brin1998anatomy}.
The empirical distribution is defined as $\hat p_2 (\cdot,v_i)  = w_{ij}/d_i$, where $w_{ij}$ is the weight of the edge and $d_i$ is the out-degree of vertex $i: d_i = \sum_{k \in N_i} w_{ik}$, where $N_i$ is the set of out-neighbors of $v_i$. In this method, for simplicity, $\lambda = d_i$. We replace $d(\cdot,\cdot)$ with KL-divergence and omit some constants, therefore:

\begin{equation}
    O_1 = - \sum_{(i,j) \in E}w_{ij}log p_2(v_i, v_j)
\end{equation}
\newline

We can embed node $v_i$ in the d-dimensional space by finding the ${ \{\overrightarrow u_i \}}_{i = 1,..|V|}$ and ${ \{\overrightarrow u'_i \}}_{i = 1,..|V|}$ in any directed/undirected graph.

In the LINE method, we embed the network with first-order and second-order proximity separately and then concatenate the embeddings by each of them for each vertex and get the desired embedding.

\subsection{Node2vec \texorpdfstring{\cite{grover2016node2vec}}{}} 

Node2vec is an embedding algorithm that maps the nodes of a knowledge graph to a low-dimensional space while preserving the network's structural properties. It starts by generating random walks on the input graph. The random walks are biased to explore both local (Breadth-First Search or BFS) and global (Depth-First Search or DFS) neighborhoods of the nodes. These walks can be considered as sentences where nodes of the graph are similar to words of a sentence. The obtained walks are fed into a skip-gram model for embedding. 
The actual reason that the developers use BFS and DFS is that they help us find similar nodes.
\newline
We measure the similarities between the embedded nodes by homophily \cite{fortunato2010community, yang2014overlapping} and structural equivalence \cite{henderson2012rolx} hypotheses. BFS emphasizes the nodes that are in the same community and follow the homophily hypothesis and DFS is used for sampling the nodes that share the same structural role and follow the structural equivalence law. Figure 2 illustrates BFS and DFS algorithms in a neighborhood.\newline
\newline
\begin{figure}
   \vspace*{1mm}
   \centering
   \includegraphics[scale=0.6]{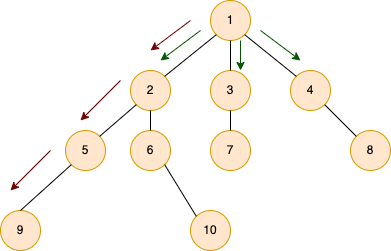}
   \caption{BFS and DFS algorithms for a neighborhood where node u is the source node \cite{grover2016node2vec}. Starting from node 1, BFS visits nodes: 1,2,3,4 and DFS visits nodes: 1,2,5,9.}
   \label{fig:figure 2}
\end{figure}

\textbf{Method}: The algorithm is based on the random walk technique which walks the graph in both the DFS and BFS fashion. 
Let’s consider $u$ as the source node, and $c_i$ as the $i^{th}$ node in the walk (therefore $c_0 = u$). The nodes in the walk are generated by this distribution:
\hfill \break
\begin{equation*}
\begin{aligned}
P(c_i = x \,|\, c_{i-1} = v) =
\begin{cases}

\frac{\pi_{vx}}{Z}  \; \; \; \; \ \ \ \ \ \ \ \ \ \ \ \ \ \left(r \in E \right) \\ \\ 

0  \ \ \ \ \ \ \ \ \ \ \ \ \ \ \ \ \ \ \ \ \ others
\end{cases}
\end{aligned}
\end{equation*}
\hfill \break

Here, $\pi$ is the unnormalized transition probability between nodes $v$ and $x$. $Z$ is the normalizing constant which is summing up all the possible values of the random variable $i$.
\hfill \break

The easiest way to bias our random walk is to sample the nodes based on the static edge weight $w_vx$ i.e. $\pi_{vx}$ = $w_vx$. In this case, if our graph is unweighted, we consider $w_vx = 1$.
While this is the simplest way, it might not be a good choice since we cannot consider network structure and explore different types of network neighborhoods. Therefore, we design our algorithm which is a 2nd order random walk (in a first-order random walk, the walker traverses the graph from one node to a randomly chosen neighbor node at each step. In the second-order random walk, the walker considers the relationships between nodes based on their neighbors before moving to the next node). Our walk depends on two parameters \textit{return parameter} or $p$ and \textit{in-out parameter} or $q$ which control how fast the walk explores and leaves the neighborhood of the starting node $u$. Let's assume the walker just traversed edge $(t,v)$ and now is at node $v$. The algorithm decides the next node based on this probability $\pi_{vx}$ = $\alpha_{pq}.w_vx$, where:

\begin{equation*}
\begin{aligned}
\alpha_{pq}.w_vx =
\begin{cases}

\frac{1}{p}  \; \; \; \; \ \ \ \ \ \ \ \ \ \ \ \ \ \ \ \ \ \ \ \ \ \ \ \ \ \ \ \ \ \ \ \ \ \left(r \in E \right) \\ \\ 

1  \ \ \ \ \ \ \ \ \ \ \ \ \ \ \ \ \ \ \ \ \ \ \ \ \ \ \ \ \ \ \ \ \ \ \ \ \ others \\ \\ 

\frac{1}{q}  \; \; \; \; \ \ \ \ \ \ \ \ \ \ \ \ \ \ \ \ \ \ \ \ \ \ \ \ \ \ \ \ \ \ \ \ \ \left(r \in E \right) \\ \\ 

\end{cases}
\end{aligned}
\end{equation*}

In this equation, $d_{tx}$ is the shortest distance between the nodes $x$ and $t$ and can be one of ${0,1,2}$ values. Based on this formula, if we set $p$ to a high value, \textit{i.e.} $(> max(q,1))$, it is less likely to revisit a node that was just visited (unless the next node in the walk has no other neighbor). This strategy leads to moderate exploration and avoids 2-hop redundancy in sampling.  
On the other hand, if we set $p$ to a low value, \textit{i.e.} $(<min(q,1))$, it is more likely that the walk is close to the source node since it leads the walk one step backward.

In addition, if we set $q>1$, the random walk is biased toward nodes close to $t$; which leads our walk to sample nodes within a small locality. On the other hand, if we set $q<1$, the walk is more likely to explore further nodes from node $t$ which encourages outward exploration.

\subsection{Predictive Text Embedding (PTE) \texorpdfstring{\cite{tang2015pte}}{}} 

Predictive Text Embedding (PTE) is an extension of the LINE method to embed heterogeneous networks. It is a semi-supervised method used for embedding text data, which means it uses both labeled and unlabeled data to train the model. The labeled and unlabeled data are represented in a large heterogeneous network and then this heterogeneous network is embedded in a low dimensional space and can be used for text embedding. Not only does this method preserve the semantic closeness of the words and documents but also it has good predictive power.
\hfill\break
Compared to the unsupervised text embedding methods such as Skip-gram or Paragraph Vectors (aka Doc2vec) \cite{le2014distributed}, which learn semantic representations of texts, the goal of this method is to learn a representation of the text that is optimized for a given text classification task. In other words, the authors anticipate the text embedding to have a strong predictive power of the performance of the given task. Since this method is applicable to different networks, we review some network definitions:
\hfill \break
\hfill \break
\textbf{Word-Word Network:} The word-word network captures the word co-occurrences in local contexts of unlabeled data. This data is the essential information used by some word embedding techniques such as skip-gram. Let $G_{ww} = (V, E_{ww})$ be a graph in which $V$ is a vocabulary of words and $E_{ww}$ is the set of edges between the words. Also, the weight $w_{ij}$ is the number of times words $v_i$ and $v_j$ appear in the context window.

\hfill \break
\textbf{Word-Document Network:} Word-document network, denoted as $G_{wd} = (V \cup D, E_{wd})$, 
is a bipartite network where $V$ is a set of words and $D$ denotes a set of documents. $E_{wd}$ is the set of edges between the words and the documents. The weight $w_{ij}$ between word $v_i$ and document $d_j$ is simply defined as the number of times $v_i$ appears in document $d_j$.
 
The mentioned networks are used for encoding unlabeled data. There is a network for encoding labeled information called the Word-Label network.

\hfill \break
\textbf{Word-Label Network:} Let’s take $G_{wd} = (V \cup L, E_{wd})$ as a bipartite network in which $V$ is the set of words, $L$ is the set of labels, and $E_{wd}$ is the set of edges connecting words and labels. The weight $w_{ij}$ of the edge between word $v_i$ and class $c_j$ is defined as $w_{ij} = \sum (d:ld=j) n_{di}$, where $n_{di}$ is the term showing frequency of word $v_i$ in document $d$, and $ld$ is the class label of document $d$. 
 
The model embeds a network that is an integration of the above networks. This type of network is called a heterogeneous text network. 

\hfill \break
\textbf{Heterogeneous Text Network:} is the combination of word-word, word-document, and word-label networks constructed from both unlabeled and labeled text data.

\hfill \break
\textbf{Method}: Given a large collection of text data with unlabeled and labeled information, the PTE algorithm tries to learn the embedding of the text by embedding the heterogeneous text network (in other words, by embedding the nodes of the heterogeneous graph) built
from the collection. 
 
 \begin{figure}
   \vspace*{1mm}
   \centering
   \includegraphics[scale=.25]{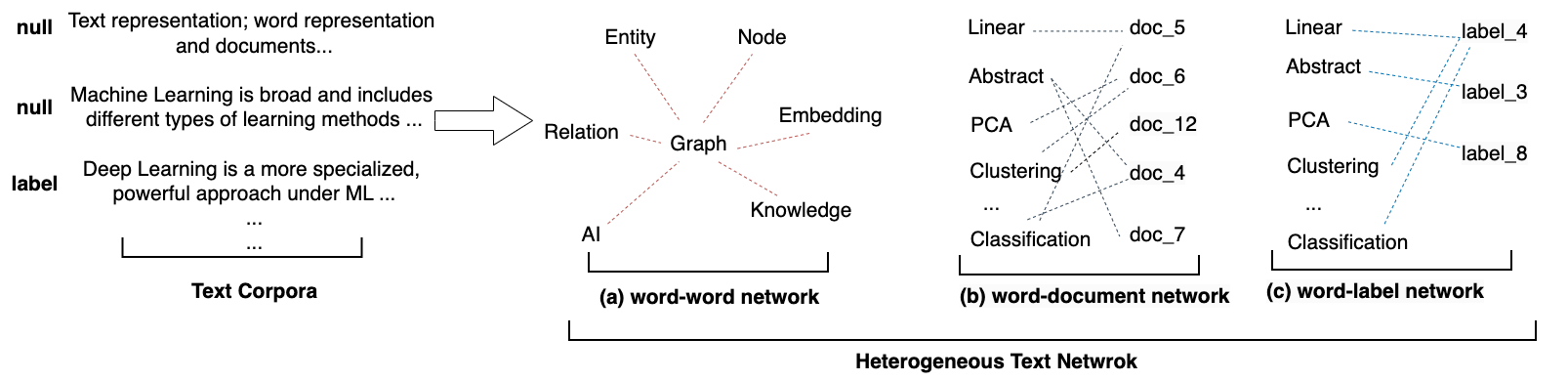}
   \caption{Converting partially labeled text corpora to a heterogeneous text network. The word-word co-occurrence and word-document networks encode the unsupervised information, capturing the local context-level and document-level word co-occurrences respectively. The word-label network encodes the supervised information, capturing the class-level word co-occurrences \cite{tang2015pte}.}
   \label{fig:figure 3}
 \end{figure}

The heterogeneous graph is made up of three different bipartite networks, and therefore we have to embed each of these graphs individually (so far there is no technique to be able to embed these graphs all together at the same time). To embed each of these bipartite graphs individually, we will use the LINE model.

As mentioned earlier, PTE is an extension of the LINE method but LINE cannot be used to embed heterogeneous networks. Therefore, to start the embedding, we use the LINE method to embed a bipartite network. To use LINE, it is essential to make use of the second-order proximity between vertices, which means that every two nodes that have similar neighbors, can be considered similar to each other which leads to closer vectors in the embedding space.  
 
Given a bipartite graph, $G = (V_A \cup V_B, E)$, $V_a$, and $V_b$ are two different types of nodes and $E$ is the set of edges between them. For every $v_i$ in $V_a$, generated by $v_j$ in $V_b$, the authors define the below formula to use the second-order proximity:

\begin{equation}
P(v_i, v_j) = \frac{\exp(u_i^T \cdot u_j)}{\sum_{i' \in A} \exp(u_{i'}^T \cdot u_j)} \quad
\end{equation}
\hfill \break

where $u_i$ is the embedding vector of vertex $v_i \in V_A$, and $u_j$ is the embedding vector of vertex $v_j \in V_B$. For each vertex $v_j \in V_B$, Equation (1) defines a conditional distribution $p(\cdot|v_j)$ over all the vertices in the set $V_A$ and for each pair of vertices $v_j, v_{j'}$, the second-order proximity is determined by their conditional distributions $p(\cdot|v_j )$, $p(\cdot|v_{j'})$ respectively. To preserve the $2_{nd}$-order proximity, we should try to make the conditional distribution $p(\cdot|v_j)$ close to its empirical distribution $p(\cdot|v_j)$. The below equation illustrates this:

\begin{equation}
O = \sum_{j \in B} \lambda_d(p(\cdot | v_j), p(\cdot | v_j)) \quad 
\end{equation}
\hfill \break
where $d(\cdot, \cdot)$ is the KL-divergence between two distributions, $\lambda_j$ is the importance of vertex $v_j$ in the network, which can be set as the degree $deg_j = \sum w_{ij}$, and the empirical distribution can be defined as $p(v_j|v_i ) = w_{ij} /deg_j$. Omitting some constants, here is the simpler version of Equation (2):

\begin{equation}
O = -\sum_{(i,j) \in E} w_{ij} \cdot \log p(v_j | v_i) \quad 
\end{equation}
\hfill \break
We can optimize the above equation using gradient descent which uses edge sampling and negative sampling.
\hfill \break 
We can embed our 3 bipartite networks using the above model. Next, we want to embed the heterogeneous text network which consists of three bipartite networks: word-word, word-document, and word-label networks. To learn the embeddings of the heterogeneous network, our approach is to collectively embed the three bipartite networks by the following equation:

\begin{equation}
O_{pte} = O_{ww} + O_{wd} + O_{wl}
\end{equation}

where $O_{ww}$, $O_{wd}$ and $O_{wl}$ are calculated individually by Equation (3). 

Once the word vectors are learned, the representation learning of any piece of text can be obtained by averaging the vectors of the words in that piece of text.

\begin{equation}
d = \frac{1}{n} \sum u_i \quad 
\end{equation}

\subsection{Metapath2vec and Metapath2vec++ \texorpdfstring{\cite{dong2017metapath2vec}}{}}

In this section, we review a neural network-based representation learning algorithm. There are recent different types of neural network-based algorithms for embedding nodes of the graphs; such as Node2vec, LINE, and DeepWalk which we discussed earlier.

Although these methods have their privileges such as the automatic discovery of latent features from the raw network, they can be applied to homogeneous networks (networks with singular types of nodes and edges). However, a large number of social and information graphs are heterogeneous, which means they have multiple types of nodes and edges. Therefore, we need new algorithms to be able to embed them. 

Here, we review metapath2vec and metapath2vec++; which are representation learning methods for embedding heterogeneous networks.
\hfill\break

\textbf{Metapath2vec}: is a representation learning method applicable to heterogeneous networks. 

The metapath2vec method develops metapath-based random walks to construct the neighborhood of a node and then feeds the achieved random walks to a skip-gram model to obtain node embeddings.
\hfill \break

\textbf{Method:} As mentioned earlier, metapath2vec generates meta path-based random walks from the nodes of the graph. The most straightforward method is to start the meta path-based walk at a random node and then move to the next random node. In this context, the probability of moving to the next node is 
$P(v_i+1|v_i)$ regardless of the node types. However, the walks are biased toward a highly visible type of nodes\cite{sun2011pathsim}. To overcome this issue, given a graph $G = (V,E,T)$ the authors design a meta path scheme to guide the walks in this form:

\begin{equation}
\rho = V_1 \xrightarrow{R_1} V_2 \xrightarrow{R_2} V_3 \xrightarrow{R_3} \ldots \ V_t\xrightarrow{R_t} V_t+1 \xrightarrow{R_{l-1}} V_l
\end{equation}
\hfill\break
Hence, the transition probability for moving to the next node is: 
\begin{equation*}
P(v^{i+1}|v^{i}_{t}, \rho) = 
\begin{aligned}
\begin{cases}
\frac{1}{|N_{t+1}(V^{i}_{t})|} \ \ \ \ \ \ \ \ \ \ \ \ \ \ \ \ \ \ \ \ \ \ \ \ \ \ \ (v^{i+1},v^{i}_{t}) \in E, 
 \phi(v^{i+1}) = t+1\\ \\ 

0  \ \ \ \ \ \ \ \ \ \ \ \ \ \ \ \ \ \ \ \ \ \ \ \ \ \ \ \ \ \ \ \ \ \ \ \ \ \ (v^{i+1},v^{i}_{t}) \in E, \phi(v^{i+1}) \neq t+1 \\ \\

0  \ \ \ \ \ \ \ \ \ \ \ \ \ \ \ \ \ \ \ \ \ \ \ \ \ \ \ \ \ \ \ \ \ \ \ \ \ \ \ (v^{i+1},v^{i}_{t}) \in E  \\ \\
\end{cases}
\end{aligned}
\end{equation*}
\textbf{}
\hfill 
 
where $v_t^i \in V_t$ and $N_{t+1} (v_t^i)$ denote the $V_{t+1}$ type of neighborhood of node $v_{t^i}$ . Also, meta-paths are designed in a symmetric way; which means the first node denoted as $V_1$ is the same as the last one, $V_l$, and therefore, has the same probability of being reached, which means:

\begin{equation}
p(v_{i+1} | v_{t}^{i} ) = p(v_{i+1} | v_{l}^{i} ), \text{ if } t = l
\end{equation}
\hfill\break
For example, consider “APA” and “APVPA” as meta path schemes where the former represents the “coauthor collaboration on a paper” and the latter represents “two authors publish papers in the same venue”.
In the next step, the algorithm inputs the achieved random walks to a heterogeneous skip-gram model to get the embeddings of the nodes. Given a heterogeneous graph $G = (V, E,T)$, the objective of using a heterogeneous skip-gram model is to maximize the network probability in terms of local structure or $N_{t(v)}, t \in T_v,$ i.e:

\begin{equation}
argmax \sum_{v \in V}\sum_{t \in T_V}\sum_{c_t \in N_t(v)} log p(c_t | v; \theta)
\end{equation}
\hfill

where $N_{t(v)}$ is $v$’s neighborhood with the $t^(th)$ type of nodes, and $p(c_t |v; \theta)$ is a softmax function, that is $p(c_t |v; \theta) = e^{X_{c_t}.X_v} /\sum(u\in V)e^{X_u.X_v}$ where $X_v$ is the $\textit{V}$th row of $X$, representing the embedding vector for node $v$. 
 
Metapath2vec recognizes the context nodes of node $v$ when constructing its neighborhood function $N_v$ based on their types. However, it ignores these types in the softmax function. Therefore, we introduce a modified version of metapath2vec to enhance the results. 
\hfill \break

\textbf{Metapath2vec++:} is an extension of metapath2vec designed to improve the quality of the embeddings in the heteregeneous graphs. In metapath2vec++, the softmax function is normalized with respect to the type of the context node $c_t$: 

\begin{equation}
p(c_t | v; \theta) = \frac{e^{X_{ct} \cdot X_v}}{\sum_{u \in V} e^{X_u \cdot X_v}} \quad (4)
\end{equation}
\hfill\break
where $p(c_t | v; \theta)$ is adjusted to the node type $t$ and $V_t$ is the node set of type $t$. In this case, we will have one set of multinomial distributions for each type of the $c_t$ in the output layer of the skip-gram model. 
Figure 4, illustrates the differences between the embedding results of some models:
\begin{figure}
   \vspace*{1mm}
   \centering
   \includegraphics[scale=.5]{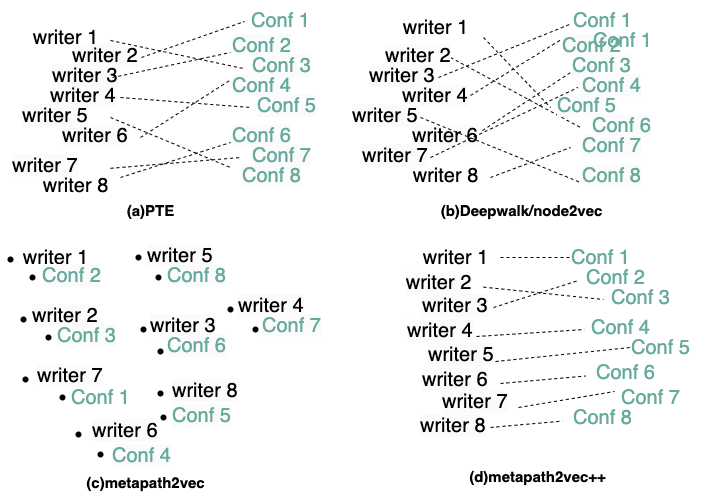}
   \caption{2D PCA projections of the 128D embeddings of 16 top CS conferences and corresponding high-profile authors \cite{dong2017metapath2vec}.}
   \label{fig:figure 4}
 \end{figure}

\subsection{Regpattern2vec \texorpdfstring{\cite{keshavarzi2021regpattern2vec}}{}}

In this section, we explain another method for representation learning of the graphs. Regpattern2vec is an embedding algorithm categorized as a deep learning method based on random walks. This algorithm uses a fixed regular pattern (regular expression) to bias the random walks and then feeds the achieved walks to a modified version of the skip-gram to embed the walks. The obtained embeddings can be used for various machine learning tasks; such as link prediction, node/edge classification, etc. In the original paper, these embeddings are used for the link prediction tasks.

Before explaining the regpattern2vec algorithm, we describe some preliminary concepts:
\newline

\textbf{Regular expression:} A regular expression or regular pattern defines a search pattern in a text in the form of a sequence of characters \cite{5392601}. In other words, a regular expression defines a set of strings that match it \cite{friedl2006mastering}. Regular expressions can contain both ordinary and special characters. Ordinary characters mean alphabets and numbers such as \textit{A,b,5}. Special characters are non-alphabets and non-numbers such as \textit{‘(‘} and \textit{‘?’}. Special characters have special meanings in the regular expressions. Some examples of regular expressions are \textit{a{3,5}},  \textit{\textasciicircum The.*Spain} , \textit{H[\textasciicircum T] + HT}. 
\newline

\textbf{Regular pattern on knowledge graph:} If $G =(V, E)$ is a knowledge graph which has a node type mapping function $\phi: V \xrightarrow T$ and an edge type mapping function $\phi: E \xrightarrow R$. A regular pattern $r$ on $G$ is formed over either set $T$ or $R$ as the alphabet. 
\hfill \break

\textbf{Finite State Machine \cite{lee1996principles}:} also called Finite State Automaton, is a mathematical representation of computation which is an abstract concept but can also be implemented in software and hardware for different purposes; such as reducing the mathematical work in the theory of computation, pattern matching and lexical analysis. A Finite State Machine can be classified into two types: Deterministic Finite Automaton (DFA) and Non-Detereminsitic Finite Automaton (NDFA/NFA). Since our model is using the DFA, we will only give a brief explanation of it here. 
As explained earlier, the codes accept a user input regex. Before running the random walk, we first want to make sure that the regex is valid and that random walks with that regex are applicable. We check this validation via DFA.
In this method, we have a DFA consisting of 5 states; in which we start from state 0 as the initial state and try to move through the states via the transition function, and then return to the final state (which is again state 0 in our case). If we arrive at the final state, this means the regex is valid; otherwise, the user has to enter a different regex. If the regex is valid, the algorithm runs the walk based on this regex. After the walk is finished, we feed the generated walks to a skip-gram and obtain the embeddings of the nodes. The generated embeddings can be used for various machine-learning tasks.
 
\hfill 

\textbf{Deterministic Finite Automata \cite{lee1996principles}:} or DFA, is a Finite State Machine that reads a string of symbols and either accepts or rejects it. For each input symbol, a state in the DFA is determined to which the machine moves. A DFA can be represented by a 5-tuple 
$(Q, \Sigma, \delta, q_{0}, F)$ in which: $Q$ denotes the set of states, $\Sigma$ (also called alphabet) denotes a finite set of symbols, $\delta$ is the transition function $(\delta: Q \times \Sigma \rightarrow Q)$, $ q_{0}$ denotes the initial state $(q_{0} \in Q)$ and $F$ denotes the final state/states $(F \subseteq Q)$.
Basically, the machine works as follows: First, it takes the string ($S$) over the alphabet ($\Sigma$) as an input. Then, starting from the initial state ($q_{0}$) while reading each character of the string $S$, the machine moves to the next state by using the transition function. If the last alphabet of $S$, makes the machine stop in $F$ (the final state/any of the final states), the machine accepts the string, otherwise, rejects it. 

\hfill \break 
\textbf{Method:} Here, we explain how regpattern2vec works. The algorithm runs on a fixed regular expression which is \textit{r = H[\textasciicircum T] + HT} based on edges of the graph. Each of \textit(H, T, \textasciicircum T) denotes an edge type and has different sub-types. The user enters a regular expression based on the edges' sub-types $(r)$ and to make sure the regular pattern is entered properly, the algorithm uses DFA to check the validation of the entered regular expression. If the user-given regular expression matches the \textit{r = H[\textasciicircum T] + HT} format and the types of the chosen edges are compatible as well, the algorithm runs the walks based on this regular expression. 
According to $r$, the random walk chooses the first edge randomly from any edges of type $H$, and the next edge is of any random edge from any type but $T$, and then chooses another random edge (from the available edges) of type $H$ and then the next edge is of type $T$. In this case, the walk length is 4, however, if the walk length is more than 4, the algorithm repeats the walk in a back-and-forth fashion. This means the walk moves backward and for the next edge, the walk chooses an edge of type $H$, and then chooses an edge of any type but $T$, then chooses type $H$ and again moves forward afterward. 
The walk repeats the same thing until it reaches the walk length.
 
At each step, the probability that a specific type of node is chosen is calculated as follows:

\begin{equation*}
\begin{aligned}
\sum^{l}_{i=1+1} P(v^{i+1}|v^{i}, M) = 
\begin{cases}
\frac{\frac{1}{N_{v_{i+1}}}}{\sum^{n}_{i=1} \frac{1}{N_v}} \ \ \ \ \ \ \ \ \ \ \ \ \ \ \ \ \ \ \ \ \ \ \ \ \ \ \ \ \ \ r \in S \\ \\ 

0  \ \ \ \ \ \ \ \ \ \ \ \ \ \ \ \ \ \ \ \ \ \ \ \ \ \ \ \ \ \ \ \ \ \ \ \ \ \ r^i,\ r^i,\ r^{i+1} \notin \ G'
\end{cases}
\end{aligned}
\end{equation*}
\textbf{}
\hfill

Here, $|N_v|$ is a degree of node $v, v_i$ indicates the current node, and $v_{i+1}$ is the next candidate node. Also, $s_i$ is the current state of $M$, and $\phi(r_i)$ is $M$’s transition function from state $s_i$ to state $s_{i+1}$.
 
The random walks based on the defined regular expression are created. Next, these random walks are fed into a modified version of a skip-gram to create the embeddings.
To capture the similarity of the edges based on their types and having close embeddings in the latent space, our modified skip-gram takes into account the types of the edges. We use these biased walks as an input to the skip-gram and the output is the embeddings of these walks.

\subsection{Subgraph2vec \texorpdfstring{\cite{bozorgi2024subgraph2vec}}{}}

Subgraph2vec is a representation learning technique that demonstrates the vector representation of the entities and relations of a knowledge graph in a low-dimensional space while maintaining their semantic meanings. 
The algorithm uses random walks and a modified version of the skip-gram to create the embeddings. In this method, the user enters a schema subgraph, with the intension to bias the random walks on a specific part of the overall knowledge graph.  A schema subgraph is a subgraph of the complete schema graph. The schema graph is in the form of an arbitrary set of integers based on the edges where each integer represents an edge. After the sub-graph is given, the algorithm chooses a random node inside the subgraph and starts the random walk. The next edge is chosen randomly inside the subgraph which moves to the next random node. The walk continues based on a parameter called \textit{walk length}. Each chosen node/edge is valid only if it is within the user-defined subgraph. This method is supposed to solve the deficiency of the previous random walk-based methods such as regpattern2vec, node2vec, and metapath2vec. In the previously mentioned methods, the walks are biased on fixed regular expressions, or sequences of note types, which are defined by experts. However, this method is based on arbitrary random walks, as long as they are within the defined subgraph. 

\hfill\break
\textbf{Method:} The user enters a schema subgraph $(s')$ in the form of integers, representing the edges in the schema subgraph. Let’s assume the user has entered this subgraph: ${s' = x_1, x_2, x_3}$, where each $x_i$ denotes an edge in the graph. After the subgraph is given, a random node is chosen within this subgraph as the starting node. The walk starts at this node and in each step of the walk, a random edge is chosen. The chosen random edge is valid only if it is within the subgraph and invalid otherwise. The probability of choosing the next edge is calculated with this formula:

\begin{equation*}
\sum^{l}_{i=1+1} P(r^{i+1}|r^{i}, S) = 
\end{equation*}

\begin{equation*}
\begin{aligned}
\begin{cases}
\frac{1}{r_{t_i}} \times \frac{1}{\sum^{n}_{i=1} t_i} \; \; \; \; \ \ \ \ \ \ \ \ \ \ \ \ \ \ \ \ \ \ \ \ \ \ \ \ \ \ \ \ \ \ \ \ \ \left(r \in S \right) \\ \\ 

0  \ \ \ \ \ \ \ \ \ \ \ \ \ \ \ \ \ \ \ \ \ \ \ \ \ \ \ \ \ \ \ \ \ \ \ \ \ \ \left(r^i,\ r^i,\ r^{i+1}\right)\notin \ G'
\end{cases}
\end{aligned}
\end{equation*}
\textbf{}
\hfill \break

where $t_i$ denotes each type of the edges connected to the current node and $r_{t_i}$ denotes the number of the edges of each type. We choose the next edge from our valid set of edges based on its probability. 
\section{Conclusion}\label{sec4}

In this work, we reviewed some methods for embedding knowledge graphs. There are five different categories of methods for embedding knowledge graphs which include: matrix factorization, generative models, deep learning, graph kernels, and edge reconstruction-based optimization models. Each of these categories includes distinct subcategories and each subcategory contains different methods. We have provided examples and its relevant model type for each subcategory in Table 2. 
\newline
In this paper, our focus is on the deep learning methods; since they have recently gained popularity due to their different benefits such as scalability, versatility, high accuracy, etc. Among deep learning methods, random walk-based ones provide a versatile and powerful tool for analyzing and modeling knowledge graphs influenced by randomness. We have summarized seven of the most important recent random walk-based algorithms for embedding knowledge graphs. In addition, we have categorized these methods based on the random walk technique they use in Table 1. 
\newline
The future direction includes summarizing non-random walk based methods of the deep learning category, methods of other categories and explaining advantages and shortcomings of these methods. We hope that this up to date survey gives better insights to the researchers in this field.  

\bibliography{sn-bibliography}


\begin{thebibliography}{85}
\ifx \bisbn   \undefined \def \bisbn  #1{ISBN #1}\fi
\ifx \binits  \undefined \def \binits#1{#1}\fi
\ifx \bauthor  \undefined \def \bauthor#1{#1}\fi
\ifx \batitle  \undefined \def \batitle#1{#1}\fi
\ifx \bjtitle  \undefined \def \bjtitle#1{#1}\fi
\ifx \bvolume  \undefined \def \bvolume#1{\textbf{#1}}\fi
\ifx \byear  \undefined \def \byear#1{#1}\fi
\ifx \bissue  \undefined \def \bissue#1{#1}\fi
\ifx \bfpage  \undefined \def \bfpage#1{#1}\fi
\ifx \blpage  \undefined \def \blpage #1{#1}\fi
\ifx \burl  \undefined \def \burl#1{\textsf{#1}}\fi
\ifx \doiurl  \undefined \def \doiurl#1{\url{https://doi.org/#1}}\fi
\ifx \betal  \undefined \def \betal{\textit{et al.}}\fi
\ifx \binstitute  \undefined \def \binstitute#1{#1}\fi
\ifx \binstitutionaled  \undefined \def \binstitutionaled#1{#1}\fi
\ifx \bctitle  \undefined \def \bctitle#1{#1}\fi
\ifx \beditor  \undefined \def \beditor#1{#1}\fi
\ifx \bpublisher  \undefined \def \bpublisher#1{#1}\fi
\ifx \bbtitle  \undefined \def \bbtitle#1{#1}\fi
\ifx \bedition  \undefined \def \bedition#1{#1}\fi
\ifx \bseriesno  \undefined \def \bseriesno#1{#1}\fi
\ifx \blocation  \undefined \def \blocation#1{#1}\fi
\ifx \bsertitle  \undefined \def \bsertitle#1{#1}\fi
\ifx \bsnm \undefined \def \bsnm#1{#1}\fi
\ifx \bsuffix \undefined \def \bsuffix#1{#1}\fi
\ifx \bparticle \undefined \def \bparticle#1{#1}\fi
\ifx \barticle \undefined \def \barticle#1{#1}\fi
\bibcommenthead
\ifx \bconfdate \undefined \def \bconfdate #1{#1}\fi
\ifx \botherref \undefined \def \botherref #1{#1}\fi
\ifx \url \undefined \def \url#1{\textsf{#1}}\fi
\ifx \bchapter \undefined \def \bchapter#1{#1}\fi
\ifx \bbook \undefined \def \bbook#1{#1}\fi
\ifx \bcomment \undefined \def \bcomment#1{#1}\fi
\ifx \oauthor \undefined \def \oauthor#1{#1}\fi
\ifx \citeauthoryear \undefined \def \citeauthoryear#1{#1}\fi
\ifx \endbibitem  \undefined \def \endbibitem {}\fi
\ifx \bconflocation  \undefined \def \bconflocation#1{#1}\fi
\ifx \arxivurl  \undefined \def \arxivurl#1{\textsf{#1}}\fi
\csname PreBibitemsHook\endcsname

\bibitem[\protect\citeauthoryear{Steiner et~al.}{2012}]{steiner2012adding}
\begin{bchapter}
\bauthor{\bsnm{Steiner}, \binits{T.}},
\bauthor{\bsnm{Verborgh}, \binits{R.}},
\bauthor{\bsnm{Troncy}, \binits{R.}},
\bauthor{\bsnm{Gabarro}, \binits{J.}},
\bauthor{\bsnm{Walle}, \binits{R.}}:
\bctitle{Adding realtime coverage to the google knowledge graph}.
In: \bbtitle{11th International Semantic Web Conference (ISWC 2012)},
vol. \bseriesno{914},
pp. \bfpage{65}--\blpage{68}
(\byear{2012}).
\bcomment{Citeseer}
\end{bchapter}
\endbibitem

\bibitem[\protect\citeauthoryear{Li et~al.}{2020}]{li2020alimekg}
\begin{bchapter}
\bauthor{\bsnm{Li}, \binits{F.-L.}},
\bauthor{\bsnm{Chen}, \binits{H.}},
\bauthor{\bsnm{Xu}, \binits{G.}},
\bauthor{\bsnm{Qiu}, \binits{T.}},
\bauthor{\bsnm{Ji}, \binits{F.}},
\bauthor{\bsnm{Zhang}, \binits{J.}},
\bauthor{\bsnm{Chen}, \binits{H.}}:
\bctitle{Alimekg: Domain knowledge graph construction and application in e-commerce}.
In: \bbtitle{Proceedings of the 29th ACM International Conference on Information \& Knowledge Management},
pp. \bfpage{2581}--\blpage{2588}
(\byear{2020})
\end{bchapter}
\endbibitem

\bibitem[\protect\citeauthoryear{Mao et~al.}{2022}]{mao2022financial}
\begin{barticle}
\bauthor{\bsnm{Mao}, \binits{X.}},
\bauthor{\bsnm{Sun}, \binits{H.}},
\bauthor{\bsnm{Zhu}, \binits{X.}},
\bauthor{\bsnm{Li}, \binits{J.}}:
\batitle{Financial fraud detection using the related-party transaction knowledge graph}.
\bjtitle{Procedia Computer Science}
\bvolume{199},
\bfpage{733}--\blpage{740}
(\byear{2022})
\end{barticle}
\endbibitem

\bibitem[\protect\citeauthoryear{Soleymani et~al.}{2023}]{soleymani2023dark}
\begin{barticle}
\bauthor{\bsnm{Soleymani}, \binits{S.}},
\bauthor{\bsnm{Gravel}, \binits{N.}},
\bauthor{\bsnm{Huang}, \binits{L.-C.}},
\bauthor{\bsnm{Yeung}, \binits{W.}},
\bauthor{\bsnm{Bozorgi}, \binits{E.}},
\bauthor{\bsnm{Bendzunas}, \binits{N.G.}},
\bauthor{\bsnm{Kochut}, \binits{K.J.}},
\bauthor{\bsnm{Kannan}, \binits{N.}}:
\batitle{Dark kinase annotation, mining, and visualization using the protein kinase ontology}.
\bjtitle{PeerJ}
\bvolume{11},
\bfpage{16087}
(\byear{2023})
\end{barticle}
\endbibitem

\bibitem[\protect\citeauthoryear{Krinkin et~al.}{2020}]{krinkin2020architecture}
\begin{bchapter}
\bauthor{\bsnm{Krinkin}, \binits{K.}},
\bauthor{\bsnm{Kulikov}, \binits{I.}},
\bauthor{\bsnm{Vodyaho}, \binits{A.}},
\bauthor{\bsnm{Zhukova}, \binits{N.}}:
\bctitle{Architecture of a telecommunications network monitoring system based on a knowledge graph}.
In: \bbtitle{2020 26th Conference of Open Innovations Association (FRUCT)},
pp. \bfpage{231}--\blpage{239}
(\byear{2020}).
\bcomment{IEEE}
\end{bchapter}
\endbibitem

\bibitem[\protect\citeauthoryear{Buchgeher et~al.}{2021}]{buchgeher2021knowledge}
\begin{barticle}
\bauthor{\bsnm{Buchgeher}, \binits{G.}},
\bauthor{\bsnm{Gabauer}, \binits{D.}},
\bauthor{\bsnm{Martinez-Gil}, \binits{J.}},
\bauthor{\bsnm{Ehrlinger}, \binits{L.}}:
\batitle{Knowledge graphs in manufacturing and production: A systematic literature review}.
\bjtitle{IEEE Access}
\bvolume{9},
\bfpage{55537}--\blpage{55554}
(\byear{2021})
\end{barticle}
\endbibitem

\bibitem[\protect\citeauthoryear{Tezerjani et~al.}{2024}]{tezerjani2024real}
\begin{botherref}
\oauthor{\bsnm{Tezerjani}, \binits{M.D.}},
\oauthor{\bsnm{Carrillo}, \binits{D.}},
\oauthor{\bsnm{Qu}, \binits{D.}},
\oauthor{\bsnm{Dhakal}, \binits{S.}},
\oauthor{\bsnm{Mirzaeinia}, \binits{A.}},
\oauthor{\bsnm{Yang}, \binits{Q.}}:
Real-time motion planning for autonomous vehicles in dynamic environments.
arXiv preprint arXiv:2406.02916
(2024)
\end{botherref}
\endbibitem

\bibitem[\protect\citeauthoryear{Ahmed et~al.}{2022}]{ahmed2022knowledge}
\begin{barticle}
\bauthor{\bsnm{Ahmed}, \binits{U.}},
\bauthor{\bsnm{Srivastava}, \binits{G.}},
\bauthor{\bsnm{Djenouri}, \binits{Y.}},
\bauthor{\bsnm{Lin}, \binits{J.C.-W.}}:
\batitle{Knowledge graph based trajectory outlier detection in sustainable smart cities}.
\bjtitle{Sustainable Cities and Society}
\bvolume{78},
\bfpage{103580}
(\byear{2022})
\end{barticle}
\endbibitem

\bibitem[\protect\citeauthoryear{Liu et~al.}{2023}]{liu2023urbankg}
\begin{barticle}
\bauthor{\bsnm{Liu}, \binits{Y.}},
\bauthor{\bsnm{Ding}, \binits{J.}},
\bauthor{\bsnm{Fu}, \binits{Y.}},
\bauthor{\bsnm{Li}, \binits{Y.}}:
\batitle{Urbankg: An urban knowledge graph system}.
\bjtitle{ACM Transactions on Intelligent Systems and Technology}
\bvolume{14}(\bissue{4}),
\bfpage{1}--\blpage{25}
(\byear{2023})
\end{barticle}
\endbibitem

\bibitem[\protect\citeauthoryear{Luo et~al.}{2003}]{luo2003spectral}
\begin{barticle}
\bauthor{\bsnm{Luo}, \binits{B.}},
\bauthor{\bsnm{Wilson}, \binits{R.C.}},
\bauthor{\bsnm{Hancock}, \binits{E.R.}}:
\batitle{Spectral embedding of graphs}.
\bjtitle{Pattern recognition}
\bvolume{36}(\bissue{10}),
\bfpage{2213}--\blpage{2230}
(\byear{2003})
\end{barticle}
\endbibitem

\bibitem[\protect\citeauthoryear{Cai et~al.}{2018}]{cai2018comprehensive}
\begin{barticle}
\bauthor{\bsnm{Cai}, \binits{H.}},
\bauthor{\bsnm{Zheng}, \binits{V.W.}},
\bauthor{\bsnm{Chang}, \binits{K.C.-C.}}:
\batitle{A comprehensive survey of graph embedding: Problems, techniques, and applications}.
\bjtitle{IEEE transactions on knowledge and data engineering}
\bvolume{30}(\bissue{9}),
\bfpage{1616}--\blpage{1637}
(\byear{2018})
\end{barticle}
\endbibitem

\bibitem[\protect\citeauthoryear{Sipser}{1996}]{sipser1996introduction}
\begin{barticle}
\bauthor{\bsnm{Sipser}, \binits{M.}}:
\batitle{Introduction to the theory of computation}.
\bjtitle{ACM Sigact News}
\bvolume{27}(\bissue{1}),
\bfpage{27}--\blpage{29}
(\byear{1996})
\end{barticle}
\endbibitem

\bibitem[\protect\citeauthoryear{Xia et~al.}{2019}]{xia2019random}
\begin{barticle}
\bauthor{\bsnm{Xia}, \binits{F.}},
\bauthor{\bsnm{Liu}, \binits{J.}},
\bauthor{\bsnm{Nie}, \binits{H.}},
\bauthor{\bsnm{Fu}, \binits{Y.}},
\bauthor{\bsnm{Wan}, \binits{L.}},
\bauthor{\bsnm{Kong}, \binits{X.}}:
\batitle{Random walks: A review of algorithms and applications}.
\bjtitle{IEEE Transactions on Emerging Topics in Computational Intelligence}
\bvolume{4}(\bissue{2}),
\bfpage{95}--\blpage{107}
(\byear{2019})
\end{barticle}
\endbibitem

\bibitem[\protect\citeauthoryear{Spielman}{2006}]{spielman2006random}
\begin{botherref}
\oauthor{\bsnm{Spielman}, \binits{D.A.}}:
Random walks on graphs
(2006)
\end{botherref}
\endbibitem

\bibitem[\protect\citeauthoryear{Tang et~al.}{2015}]{tang2015line}
\begin{bchapter}
\bauthor{\bsnm{Tang}, \binits{J.}},
\bauthor{\bsnm{Qu}, \binits{M.}},
\bauthor{\bsnm{Wang}, \binits{M.}},
\bauthor{\bsnm{Zhang}, \binits{M.}},
\bauthor{\bsnm{Yan}, \binits{J.}},
\bauthor{\bsnm{Mei}, \binits{Q.}}:
\bctitle{Line: Large-scale information network embedding}.
In: \bbtitle{Proceedings of the 24th International Conference on World Wide Web},
pp. \bfpage{1067}--\blpage{1077}
(\byear{2015})
\end{bchapter}
\endbibitem

\bibitem[\protect\citeauthoryear{Wu et~al.}{2020}]{wu2020toward}
\begin{barticle}
\bauthor{\bsnm{Wu}, \binits{J.}},
\bauthor{\bsnm{Wen}, \binits{M.}},
\bauthor{\bsnm{Lu}, \binits{R.}},
\bauthor{\bsnm{Li}, \binits{B.}},
\bauthor{\bsnm{Li}, \binits{J.}}:
\batitle{Toward efficient and effective bullying detection in online social network}.
\bjtitle{Peer-to-Peer networking and Applications}
\bvolume{13}(\bissue{5}),
\bfpage{1567}--\blpage{1576}
(\byear{2020})
\end{barticle}
\endbibitem

\bibitem[\protect\citeauthoryear{Li et~al.}{2018}]{li2018link}
\begin{barticle}
\bauthor{\bsnm{Li}, \binits{J.-c.}},
\bauthor{\bsnm{Zhao}, \binits{D.-l.}},
\bauthor{\bsnm{Ge}, \binits{B.-F.}},
\bauthor{\bsnm{Yang}, \binits{K.-W.}},
\bauthor{\bsnm{Chen}, \binits{Y.-W.}}:
\batitle{A link prediction method for heterogeneous networks based on bp neural network}.
\bjtitle{Physica A: Statistical Mechanics and its Applications}
\bvolume{495},
\bfpage{1}--\blpage{17}
(\byear{2018})
\end{barticle}
\endbibitem

\bibitem[\protect\citeauthoryear{Wang et~al.}{2022}]{wang2022survey}
\begin{barticle}
\bauthor{\bsnm{Wang}, \binits{X.}},
\bauthor{\bsnm{Bo}, \binits{D.}},
\bauthor{\bsnm{Shi}, \binits{C.}},
\bauthor{\bsnm{Fan}, \binits{S.}},
\bauthor{\bsnm{Ye}, \binits{Y.}},
\bauthor{\bsnm{Philip}, \binits{S.Y.}}:
\batitle{A survey on heterogeneous graph embedding: methods, techniques, applications and sources}.
\bjtitle{IEEE Transactions on Big Data}
\bvolume{9}(\bissue{2}),
\bfpage{415}--\blpage{436}
(\byear{2022})
\end{barticle}
\endbibitem

\bibitem[\protect\citeauthoryear{Perozzi et~al.}{2014}]{perozzi2014deepwalk}
\begin{bchapter}
\bauthor{\bsnm{Perozzi}, \binits{B.}},
\bauthor{\bsnm{Al-Rfou}, \binits{R.}},
\bauthor{\bsnm{Skiena}, \binits{S.}}:
\bctitle{Deepwalk: Online learning of social representations}.
In: \bbtitle{Proceedings of the 20th ACM SIGKDD International Conference on Knowledge Discovery and Data Mining},
pp. \bfpage{701}--\blpage{710}
(\byear{2014})
\end{bchapter}
\endbibitem

\bibitem[\protect\citeauthoryear{Tang and Liu}{2009}]{tang2009relational}
\begin{bchapter}
\bauthor{\bsnm{Tang}, \binits{L.}},
\bauthor{\bsnm{Liu}, \binits{H.}}:
\bctitle{Relational learning via latent social dimensions}.
In: \bbtitle{Proceedings of the 15th ACM SIGKDD International Conference on Knowledge Discovery and Data Mining},
pp. \bfpage{817}--\blpage{826}
(\byear{2009})
\end{bchapter}
\endbibitem

\bibitem[\protect\citeauthoryear{Tang and Liu}{2011}]{tang2011leveraging}
\begin{barticle}
\bauthor{\bsnm{Tang}, \binits{L.}},
\bauthor{\bsnm{Liu}, \binits{H.}}:
\batitle{Leveraging social media networks for classification}.
\bjtitle{Data mining and knowledge discovery}
\bvolume{23},
\bfpage{447}--\blpage{478}
(\byear{2011})
\end{barticle}
\endbibitem

\bibitem[\protect\citeauthoryear{Church}{2017}]{church2017word2vec}
\begin{barticle}
\bauthor{\bsnm{Church}, \binits{K.W.}}:
\batitle{Word2vec}.
\bjtitle{Natural Language Engineering}
\bvolume{23}(\bissue{1}),
\bfpage{155}--\blpage{162}
(\byear{2017})
\end{barticle}
\endbibitem

\bibitem[\protect\citeauthoryear{Tenenbaum et~al.}{2000}]{tenenbaum2000global}
\begin{barticle}
\bauthor{\bsnm{Tenenbaum}, \binits{J.B.}},
\bauthor{\bsnm{Silva}, \binits{V.d.}},
\bauthor{\bsnm{Langford}, \binits{J.C.}}:
\batitle{A global geometric framework for nonlinear dimensionality reduction}.
\bjtitle{science}
\bvolume{290}(\bissue{5500}),
\bfpage{2319}--\blpage{2323}
(\byear{2000})
\end{barticle}
\endbibitem

\bibitem[\protect\citeauthoryear{Lin et~al.}{2005}]{lin2005semantic}
\begin{bchapter}
\bauthor{\bsnm{Lin}, \binits{Y.-Y.}},
\bauthor{\bsnm{Liu}, \binits{T.-L.}},
\bauthor{\bsnm{Chen}, \binits{H.-T.}}:
\bctitle{Semantic manifold learning for image retrieval}.
In: \bbtitle{Proceedings of the 13th Annual ACM International Conference on Multimedia},
pp. \bfpage{249}--\blpage{258}
(\byear{2005})
\end{bchapter}
\endbibitem

\bibitem[\protect\citeauthoryear{He and Niyogi}{2003}]{he2003locality}
\begin{botherref}
\oauthor{\bsnm{He}, \binits{X.}},
\oauthor{\bsnm{Niyogi}, \binits{P.}}:
Locality preserving projections.
Advances in neural information processing systems
\textbf{16}
(2003)
\end{botherref}
\endbibitem

\bibitem[\protect\citeauthoryear{Gong et~al.}{2014}]{gong2014signed}
\begin{bchapter}
\bauthor{\bsnm{Gong}, \binits{C.}},
\bauthor{\bsnm{Tao}, \binits{D.}},
\bauthor{\bsnm{Yang}, \binits{J.}},
\bauthor{\bsnm{Fu}, \binits{K.}}:
\bctitle{Signed laplacian embedding for supervised dimension reduction}.
In: \bbtitle{Proceedings of the AAAI Conference on Artificial Intelligence},
vol. \bseriesno{28}
(\byear{2014})
\end{bchapter}
\endbibitem

\bibitem[\protect\citeauthoryear{Sun et~al.}{2008}]{sun2008hypergraph}
\begin{bchapter}
\bauthor{\bsnm{Sun}, \binits{L.}},
\bauthor{\bsnm{Ji}, \binits{S.}},
\bauthor{\bsnm{Ye}, \binits{J.}}:
\bctitle{Hypergraph spectral learning for multi-label classification}.
In: \bbtitle{Proceedings of the 14th ACM SIGKDD International Conference on Knowledge Discovery and Data Mining},
pp. \bfpage{668}--\blpage{676}
(\byear{2008})
\end{bchapter}
\endbibitem

\bibitem[\protect\citeauthoryear{Roweis and Saul}{2000}]{roweis2000nonlinear}
\begin{barticle}
\bauthor{\bsnm{Roweis}, \binits{S.T.}},
\bauthor{\bsnm{Saul}, \binits{L.K.}}:
\batitle{Nonlinear dimensionality reduction by locally linear embedding}.
\bjtitle{science}
\bvolume{290}(\bissue{5500}),
\bfpage{2323}--\blpage{2326}
(\byear{2000})
\end{barticle}
\endbibitem

\bibitem[\protect\citeauthoryear{Cao et~al.}{2015}]{cao2015grarep}
\begin{bchapter}
\bauthor{\bsnm{Cao}, \binits{S.}},
\bauthor{\bsnm{Lu}, \binits{W.}},
\bauthor{\bsnm{Xu}, \binits{Q.}}:
\bctitle{Grarep: Learning graph representations with global structural information}.
In: \bbtitle{Proceedings of the 24th ACM International on Conference on Information and Knowledge Management},
pp. \bfpage{891}--\blpage{900}
(\byear{2015})
\end{bchapter}
\endbibitem

\bibitem[\protect\citeauthoryear{Nie et~al.}{2017}]{nie2017unsupervised}
\begin{bchapter}
\bauthor{\bsnm{Nie}, \binits{F.}},
\bauthor{\bsnm{Zhu}, \binits{W.}},
\bauthor{\bsnm{Li}, \binits{X.}}:
\bctitle{Unsupervised large graph embedding}.
In: \bbtitle{Proceedings of the AAAI Conference on Artificial Intelligence},
vol. \bseriesno{31}
(\byear{2017})
\end{bchapter}
\endbibitem

\bibitem[\protect\citeauthoryear{Pang et~al.}{2017}]{pang2017flexible}
\begin{bchapter}
\bauthor{\bsnm{Pang}, \binits{T.}},
\bauthor{\bsnm{Nie}, \binits{F.}},
\bauthor{\bsnm{Han}, \binits{J.}}:
\bctitle{Flexible orthogonal neighborhood preserving embedding.}
In: \bbtitle{IJCAI},
vol. \bseriesno{361},
pp. \bfpage{2592}--\blpage{2598}
(\byear{2017})
\end{bchapter}
\endbibitem

\bibitem[\protect\citeauthoryear{Shaw and Jebara}{2009}]{shaw2009structure}
\begin{bchapter}
\bauthor{\bsnm{Shaw}, \binits{B.}},
\bauthor{\bsnm{Jebara}, \binits{T.}}:
\bctitle{Structure preserving embedding}.
In: \bbtitle{Proceedings of the 26th Annual International Conference on Machine Learning},
pp. \bfpage{937}--\blpage{944}
(\byear{2009})
\end{bchapter}
\endbibitem

\bibitem[\protect\citeauthoryear{Grover and Leskovec}{2016}]{grover2016node2vec}
\begin{bchapter}
\bauthor{\bsnm{Grover}, \binits{A.}},
\bauthor{\bsnm{Leskovec}, \binits{J.}}:
\bctitle{node2vec: Scalable feature learning for networks}.
In: \bbtitle{Proceedings of the 22nd ACM SIGKDD International Conference on Knowledge Discovery and Data Mining},
pp. \bfpage{855}--\blpage{864}
(\byear{2016})
\end{bchapter}
\endbibitem

\bibitem[\protect\citeauthoryear{Tang et~al.}{2015}]{tang2015pte}
\begin{bchapter}
\bauthor{\bsnm{Tang}, \binits{J.}},
\bauthor{\bsnm{Qu}, \binits{M.}},
\bauthor{\bsnm{Mei}, \binits{Q.}}:
\bctitle{Pte: Predictive text embedding through large-scale heterogeneous text networks}.
In: \bbtitle{Proceedings of the 21th ACM SIGKDD International Conference on Knowledge Discovery and Data Mining},
pp. \bfpage{1165}--\blpage{1174}
(\byear{2015})
\end{bchapter}
\endbibitem

\bibitem[\protect\citeauthoryear{Dong et~al.}{2017}]{dong2017metapath2vec}
\begin{bchapter}
\bauthor{\bsnm{Dong}, \binits{Y.}},
\bauthor{\bsnm{Chawla}, \binits{N.V.}},
\bauthor{\bsnm{Swami}, \binits{A.}}:
\bctitle{metapath2vec: Scalable representation learning for heterogeneous networks}.
In: \bbtitle{Proceedings of the 23rd ACM SIGKDD International Conference on Knowledge Discovery and Data Mining},
pp. \bfpage{135}--\blpage{144}
(\byear{2017})
\end{bchapter}
\endbibitem

\bibitem[\protect\citeauthoryear{Bozorgi et~al.}{2024}]{bozorgi2024subgraph2vec}
\begin{botherref}
\oauthor{\bsnm{Bozorgi}, \binits{E.}},
\oauthor{\bsnm{Soleimani}, \binits{S.}},
\oauthor{\bsnm{Alqaiidi}, \binits{S.K.}},
\oauthor{\bsnm{Arabnia}, \binits{H.R.}},
\oauthor{\bsnm{Kochut}, \binits{K.}}:
Subgraph2vec: A random walk-based algorithm for embedding knowledge graphs.
arXiv preprint arXiv:2405.02240
(2024)
\end{botherref}
\endbibitem

\bibitem[\protect\citeauthoryear{Keshavarzi et~al.}{2021}]{keshavarzi2021regpattern2vec}
\begin{bchapter}
\bauthor{\bsnm{Keshavarzi}, \binits{A.}},
\bauthor{\bsnm{Kannan}, \binits{N.}},
\bauthor{\bsnm{Kochut}, \binits{K.}}:
\bctitle{Regpattern2vec: link prediction in knowledge graphs}.
In: \bbtitle{2021 IEEE International IOT, Electronics and Mechatronics Conference (IEMTRONICS)},
pp. \bfpage{1}--\blpage{7}
(\byear{2021}).
\bcomment{IEEE}
\end{bchapter}
\endbibitem

\bibitem[\protect\citeauthoryear{Bruna et~al.}{2013}]{bruna2013spectral}
\begin{botherref}
\oauthor{\bsnm{Bruna}, \binits{J.}},
\oauthor{\bsnm{Zaremba}, \binits{W.}},
\oauthor{\bsnm{Szlam}, \binits{A.}},
\oauthor{\bsnm{LeCun}, \binits{Y.}}:
Spectral networks and locally connected networks on graphs.
arXiv preprint arXiv:1312.6203
(2013)
\end{botherref}
\endbibitem

\bibitem[\protect\citeauthoryear{Monti et~al.}{2017}]{monti2017geometric}
\begin{bchapter}
\bauthor{\bsnm{Monti}, \binits{F.}},
\bauthor{\bsnm{Boscaini}, \binits{D.}},
\bauthor{\bsnm{Masci}, \binits{J.}},
\bauthor{\bsnm{Rodola}, \binits{E.}},
\bauthor{\bsnm{Svoboda}, \binits{J.}},
\bauthor{\bsnm{Bronstein}, \binits{M.M.}}:
\bctitle{Geometric deep learning on graphs and manifolds using mixture model cnns}.
In: \bbtitle{Proceedings of the IEEE Conference on Computer Vision and Pattern Recognition},
pp. \bfpage{5115}--\blpage{5124}
(\byear{2017})
\end{bchapter}
\endbibitem

\bibitem[\protect\citeauthoryear{Wang et~al.}{2016}]{wang2016structural}
\begin{bchapter}
\bauthor{\bsnm{Wang}, \binits{D.}},
\bauthor{\bsnm{Cui}, \binits{P.}},
\bauthor{\bsnm{Zhu}, \binits{W.}}:
\bctitle{Structural deep network embedding}.
In: \bbtitle{Proceedings of the 22nd ACM SIGKDD International Conference on Knowledge Discovery and Data Mining},
pp. \bfpage{1225}--\blpage{1234}
(\byear{2016})
\end{bchapter}
\endbibitem

\bibitem[\protect\citeauthoryear{Scarselli et~al.}{2008}]{scarselli2008graph}
\begin{barticle}
\bauthor{\bsnm{Scarselli}, \binits{F.}},
\bauthor{\bsnm{Gori}, \binits{M.}},
\bauthor{\bsnm{Tsoi}, \binits{A.C.}},
\bauthor{\bsnm{Hagenbuchner}, \binits{M.}},
\bauthor{\bsnm{Monfardini}, \binits{G.}}:
\batitle{The graph neural network model}.
\bjtitle{IEEE transactions on neural networks}
\bvolume{20}(\bissue{1}),
\bfpage{61}--\blpage{80}
(\byear{2008})
\end{barticle}
\endbibitem

\bibitem[\protect\citeauthoryear{Geng et~al.}{2015}]{geng2015learning}
\begin{bchapter}
\bauthor{\bsnm{Geng}, \binits{X.}},
\bauthor{\bsnm{Zhang}, \binits{H.}},
\bauthor{\bsnm{Bian}, \binits{J.}},
\bauthor{\bsnm{Chua}, \binits{T.-S.}}:
\bctitle{Learning image and user features for recommendation in social networks}.
In: \bbtitle{Proceedings of the IEEE International Conference on Computer Vision},
pp. \bfpage{4274}--\blpage{4282}
(\byear{2015})
\end{bchapter}
\endbibitem

\bibitem[\protect\citeauthoryear{Man et~al.}{2016}]{man2016predict}
\begin{bchapter}
\bauthor{\bsnm{Man}, \binits{T.}},
\bauthor{\bsnm{Shen}, \binits{H.}},
\bauthor{\bsnm{Liu}, \binits{S.}},
\bauthor{\bsnm{Jin}, \binits{X.}},
\bauthor{\bsnm{Cheng}, \binits{X.}}:
\bctitle{Predict anchor links across social networks via an embedding approach.}
In: \bbtitle{Ijcai},
vol. \bseriesno{16},
pp. \bfpage{1823}--\blpage{1829}
(\byear{2016})
\end{bchapter}
\endbibitem

\bibitem[\protect\citeauthoryear{Zhou et~al.}{2017}]{zhou2017scalable}
\begin{bchapter}
\bauthor{\bsnm{Zhou}, \binits{C.}},
\bauthor{\bsnm{Liu}, \binits{Y.}},
\bauthor{\bsnm{Liu}, \binits{X.}},
\bauthor{\bsnm{Liu}, \binits{Z.}},
\bauthor{\bsnm{Gao}, \binits{J.}}:
\bctitle{Scalable graph embedding for asymmetric proximity}.
In: \bbtitle{Proceedings of the AAAI Conference on Artificial Intelligence},
vol. \bseriesno{31}
(\byear{2017})
\end{bchapter}
\endbibitem

\bibitem[\protect\citeauthoryear{Xiong et~al.}{2017}]{xiong2017explicit}
\begin{bchapter}
\bauthor{\bsnm{Xiong}, \binits{C.}},
\bauthor{\bsnm{Power}, \binits{R.}},
\bauthor{\bsnm{Callan}, \binits{J.}}:
\bctitle{Explicit semantic ranking for academic search via knowledge graph embedding}.
In: \bbtitle{Proceedings of the 26th International Conference on World Wide Web},
pp. \bfpage{1271}--\blpage{1279}
(\byear{2017})
\end{bchapter}
\endbibitem

\bibitem[\protect\citeauthoryear{Feng et~al.}{2016}]{feng2016gake}
\begin{bchapter}
\bauthor{\bsnm{Feng}, \binits{J.}},
\bauthor{\bsnm{Huang}, \binits{M.}},
\bauthor{\bsnm{Yang}, \binits{Y.}},
\bauthor{\bsnm{Zhu}, \binits{X.}}:
\bctitle{Gake: Graph aware knowledge embedding}.
In: \bbtitle{Proceedings of COLING 2016, the 26th International Conference on Computational Linguistics: Technical Papers},
pp. \bfpage{641}--\blpage{651}
(\byear{2016})
\end{bchapter}
\endbibitem

\bibitem[\protect\citeauthoryear{Ren et~al.}{2016}]{ren2016label}
\begin{bchapter}
\bauthor{\bsnm{Ren}, \binits{X.}},
\bauthor{\bsnm{He}, \binits{W.}},
\bauthor{\bsnm{Qu}, \binits{M.}},
\bauthor{\bsnm{Voss}, \binits{C.R.}},
\bauthor{\bsnm{Ji}, \binits{H.}},
\bauthor{\bsnm{Han}, \binits{J.}}:
\bctitle{Label noise reduction in entity typing by heterogeneous partial-label embedding}.
In: \bbtitle{Proceedings of the 22nd ACM SIGKDD International Conference on Knowledge Discovery and Data Mining},
pp. \bfpage{1825}--\blpage{1834}
(\byear{2016})
\end{bchapter}
\endbibitem

\bibitem[\protect\citeauthoryear{Gui et~al.}{2016}]{gui2016large}
\begin{bchapter}
\bauthor{\bsnm{Gui}, \binits{H.}},
\bauthor{\bsnm{Liu}, \binits{J.}},
\bauthor{\bsnm{Tao}, \binits{F.}},
\bauthor{\bsnm{Jiang}, \binits{M.}},
\bauthor{\bsnm{Norick}, \binits{B.}},
\bauthor{\bsnm{Han}, \binits{J.}}:
\bctitle{Large-scale embedding learning in heterogeneous event data}.
In: \bbtitle{2016 IEEE 16th International Conference on Data Mining (ICDM)},
pp. \bfpage{907}--\blpage{912}
(\byear{2016}).
\bcomment{IEEE}
\end{bchapter}
\endbibitem

\bibitem[\protect\citeauthoryear{Liu et~al.}{2016}]{liu2016aligning}
\begin{bchapter}
\bauthor{\bsnm{Liu}, \binits{L.}},
\bauthor{\bsnm{Cheung}, \binits{W.K.}},
\bauthor{\bsnm{Li}, \binits{X.}},
\bauthor{\bsnm{Liao}, \binits{L.}}:
\bctitle{Aligning users across social networks using network embedding.}
In: \bbtitle{Ijcai},
vol. \bseriesno{16},
pp. \bfpage{1774}--\blpage{80}
(\byear{2016})
\end{bchapter}
\endbibitem

\bibitem[\protect\citeauthoryear{Zhang et~al.}{2017}]{zhang2017regions}
\begin{bchapter}
\bauthor{\bsnm{Zhang}, \binits{C.}},
\bauthor{\bsnm{Zhang}, \binits{K.}},
\bauthor{\bsnm{Yuan}, \binits{Q.}},
\bauthor{\bsnm{Peng}, \binits{H.}},
\bauthor{\bsnm{Zheng}, \binits{Y.}},
\bauthor{\bsnm{Hanratty}, \binits{T.}},
\bauthor{\bsnm{Wang}, \binits{S.}},
\bauthor{\bsnm{Han}, \binits{J.}}:
\bctitle{Regions, periods, activities: Uncovering urban dynamics via cross-modal representation learning}.
In: \bbtitle{Proceedings of the 26th International Conference on World Wide Web},
pp. \bfpage{361}--\blpage{370}
(\byear{2017})
\end{bchapter}
\endbibitem

\bibitem[\protect\citeauthoryear{Bordes et~al.}{2013}]{bordes2013translating}
\begin{botherref}
\oauthor{\bsnm{Bordes}, \binits{A.}},
\oauthor{\bsnm{Usunier}, \binits{N.}},
\oauthor{\bsnm{Garcia-Duran}, \binits{A.}},
\oauthor{\bsnm{Weston}, \binits{J.}},
\oauthor{\bsnm{Yakhnenko}, \binits{O.}}:
Translating embeddings for modeling multi-relational data.
Advances in neural information processing systems
\textbf{26}
(2013)
\end{botherref}
\endbibitem

\bibitem[\protect\citeauthoryear{Wang et~al.}{2014}]{wang2014knowledge}
\begin{bchapter}
\bauthor{\bsnm{Wang}, \binits{Z.}},
\bauthor{\bsnm{Zhang}, \binits{J.}},
\bauthor{\bsnm{Feng}, \binits{J.}},
\bauthor{\bsnm{Chen}, \binits{Z.}}:
\bctitle{Knowledge graph embedding by translating on hyperplanes}.
In: \bbtitle{Proceedings of the AAAI Conference on Artificial Intelligence},
vol. \bseriesno{28}
(\byear{2014})
\end{bchapter}
\endbibitem

\bibitem[\protect\citeauthoryear{Lin et~al.}{2015}]{lin2015learning}
\begin{bchapter}
\bauthor{\bsnm{Lin}, \binits{Y.}},
\bauthor{\bsnm{Liu}, \binits{Z.}},
\bauthor{\bsnm{Sun}, \binits{M.}},
\bauthor{\bsnm{Liu}, \binits{Y.}},
\bauthor{\bsnm{Zhu}, \binits{X.}}:
\bctitle{Learning entity and relation embeddings for knowledge graph completion}.
In: \bbtitle{Proceedings of the AAAI Conference on Artificial Intelligence},
vol. \bseriesno{29}
(\byear{2015})
\end{bchapter}
\endbibitem

\bibitem[\protect\citeauthoryear{Ji et~al.}{2015}]{ji2015knowledge}
\begin{bchapter}
\bauthor{\bsnm{Ji}, \binits{G.}},
\bauthor{\bsnm{He}, \binits{S.}},
\bauthor{\bsnm{Xu}, \binits{L.}},
\bauthor{\bsnm{Liu}, \binits{K.}},
\bauthor{\bsnm{Zhao}, \binits{J.}}:
\bctitle{Knowledge graph embedding via dynamic mapping matrix}.
In: \bbtitle{Proceedings of the 53rd Annual Meeting of the Association for Computational Linguistics and the 7th International Joint Conference on Natural Language Processing (volume 1: Long Papers)},
pp. \bfpage{687}--\blpage{696}
(\byear{2015})
\end{bchapter}
\endbibitem

\bibitem[\protect\citeauthoryear{Socher et~al.}{2013}]{socher2013reasoning}
\begin{botherref}
\oauthor{\bsnm{Socher}, \binits{R.}},
\oauthor{\bsnm{Chen}, \binits{D.}},
\oauthor{\bsnm{Manning}, \binits{C.D.}},
\oauthor{\bsnm{Ng}, \binits{A.}}:
Reasoning with neural tensor networks for knowledge base completion.
Advances in neural information processing systems
\textbf{26}
(2013)
\end{botherref}
\endbibitem

\bibitem[\protect\citeauthoryear{Yang et~al.}{2014}]{yang2014embedding}
\begin{botherref}
\oauthor{\bsnm{Yang}, \binits{B.}},
\oauthor{\bsnm{Yih}, \binits{W.-t.}},
\oauthor{\bsnm{He}, \binits{X.}},
\oauthor{\bsnm{Gao}, \binits{J.}},
\oauthor{\bsnm{Deng}, \binits{L.}}:
Embedding entities and relations for learning and inference in knowledge bases.
arXiv preprint arXiv:1412.6575
(2014)
\end{botherref}
\endbibitem

\bibitem[\protect\citeauthoryear{Trouillon et~al.}{2016}]{trouillon2016complex}
\begin{bchapter}
\bauthor{\bsnm{Trouillon}, \binits{T.}},
\bauthor{\bsnm{Welbl}, \binits{J.}},
\bauthor{\bsnm{Riedel}, \binits{S.}},
\bauthor{\bsnm{Gaussier}, \binits{{\'E}.}},
\bauthor{\bsnm{Bouchard}, \binits{G.}}:
\bctitle{Complex embeddings for simple link prediction}.
In: \bbtitle{International Conference on Machine Learning},
pp. \bfpage{2071}--\blpage{2080}
(\byear{2016}).
\bcomment{PMLR}
\end{bchapter}
\endbibitem

\bibitem[\protect\citeauthoryear{Sun et~al.}{2019}]{sun2019rotate}
\begin{botherref}
\oauthor{\bsnm{Sun}, \binits{Z.}},
\oauthor{\bsnm{Deng}, \binits{Z.-H.}},
\oauthor{\bsnm{Nie}, \binits{J.-Y.}},
\oauthor{\bsnm{Tang}, \binits{J.}}:
Rotate: Knowledge graph embedding by relational rotation in complex space.
arXiv preprint arXiv:1902.10197
(2019)
\end{botherref}
\endbibitem

\bibitem[\protect\citeauthoryear{Pr{\v{z}}ulj}{2007}]{prvzulj2007biological}
\begin{barticle}
\bauthor{\bsnm{Pr{\v{z}}ulj}, \binits{N.}}:
\batitle{Biological network comparison using graphlet degree distribution}.
\bjtitle{Bioinformatics}
\bvolume{23}(\bissue{2}),
\bfpage{177}--\blpage{183}
(\byear{2007})
\end{barticle}
\endbibitem

\bibitem[\protect\citeauthoryear{Tong and Faloutsos}{2006}]{tong2006center}
\begin{bchapter}
\bauthor{\bsnm{Tong}, \binits{H.}},
\bauthor{\bsnm{Faloutsos}, \binits{C.}}:
\bctitle{Center-piece subgraphs: problem definition and fast solutions}.
In: \bbtitle{Proceedings of the 12th ACM SIGKDD International Conference on Knowledge Discovery and Data Mining},
pp. \bfpage{404}--\blpage{413}
(\byear{2006})
\end{bchapter}
\endbibitem

\bibitem[\protect\citeauthoryear{Narayanan et~al.}{2017}]{narayanan2017graph2vec}
\begin{botherref}
\oauthor{\bsnm{Narayanan}, \binits{A.}},
\oauthor{\bsnm{Chandramohan}, \binits{M.}},
\oauthor{\bsnm{Venkatesan}, \binits{R.}},
\oauthor{\bsnm{Chen}, \binits{L.}},
\oauthor{\bsnm{Liu}, \binits{Y.}},
\oauthor{\bsnm{Jaiswal}, \binits{S.}}:
graph2vec: Learning distributed representations of graphs.
arXiv preprint arXiv:1707.05005
(2017)
\end{botherref}
\endbibitem

\bibitem[\protect\citeauthoryear{Cao et~al.}{2016}]{cao2016deep}
\begin{bchapter}
\bauthor{\bsnm{Cao}, \binits{S.}},
\bauthor{\bsnm{Lu}, \binits{W.}},
\bauthor{\bsnm{Xu}, \binits{Q.}}:
\bctitle{Deep neural networks for learning graph representations}.
In: \bbtitle{Proceedings of the AAAI Conference on Artificial Intelligence},
vol. \bseriesno{30}
(\byear{2016})
\end{bchapter}
\endbibitem

\bibitem[\protect\citeauthoryear{Shervashidze et~al.}{2009}]{shervashidze2009efficient}
\begin{bchapter}
\bauthor{\bsnm{Shervashidze}, \binits{N.}},
\bauthor{\bsnm{Vishwanathan}, \binits{S.}},
\bauthor{\bsnm{Petri}, \binits{T.}},
\bauthor{\bsnm{Mehlhorn}, \binits{K.}},
\bauthor{\bsnm{Borgwardt}, \binits{K.}}:
\bctitle{Efficient graphlet kernels for large graph comparison}.
In: \bbtitle{Artificial Intelligence and Statistics},
pp. \bfpage{488}--\blpage{495}
(\byear{2009}).
\bcomment{PMLR}
\end{bchapter}
\endbibitem

\bibitem[\protect\citeauthoryear{Ying et~al.}{2018}]{ying2018hierarchical}
\begin{botherref}
\oauthor{\bsnm{Ying}, \binits{Z.}},
\oauthor{\bsnm{You}, \binits{J.}},
\oauthor{\bsnm{Morris}, \binits{C.}},
\oauthor{\bsnm{Ren}, \binits{X.}},
\oauthor{\bsnm{Hamilton}, \binits{W.}},
\oauthor{\bsnm{Leskovec}, \binits{J.}}:
Hierarchical graph representation learning with differentiable pooling.
Advances in neural information processing systems
\textbf{31}
(2018)
\end{botherref}
\endbibitem

\bibitem[\protect\citeauthoryear{Xu et~al.}{2018}]{xu2018powerful}
\begin{botherref}
\oauthor{\bsnm{Xu}, \binits{K.}},
\oauthor{\bsnm{Hu}, \binits{W.}},
\oauthor{\bsnm{Leskovec}, \binits{J.}},
\oauthor{\bsnm{Jegelka}, \binits{S.}}:
How powerful are graph neural networks?
arXiv preprint arXiv:1810.00826
(2018)
\end{botherref}
\endbibitem

\bibitem[\protect\citeauthoryear{Alsentzer et~al.}{2020}]{alsentzer2020subgraph}
\begin{barticle}
\bauthor{\bsnm{Alsentzer}, \binits{E.}},
\bauthor{\bsnm{Finlayson}, \binits{S.}},
\bauthor{\bsnm{Li}, \binits{M.}},
\bauthor{\bsnm{Zitnik}, \binits{M.}}:
\batitle{Subgraph neural networks}.
\bjtitle{Advances in Neural Information Processing Systems}
\bvolume{33},
\bfpage{8017}--\blpage{8029}
(\byear{2020})
\end{barticle}
\endbibitem

\bibitem[\protect\citeauthoryear{Shervashidze et~al.}{2011}]{shervashidze2011weisfeiler}
\begin{botherref}
\oauthor{\bsnm{Shervashidze}, \binits{N.}},
\oauthor{\bsnm{Schweitzer}, \binits{P.}},
\oauthor{\bsnm{Van~Leeuwen}, \binits{E.J.}},
\oauthor{\bsnm{Mehlhorn}, \binits{K.}},
\oauthor{\bsnm{Borgwardt}, \binits{K.M.}}:
Weisfeiler-lehman graph kernels.
Journal of Machine Learning Research
\textbf{12}(9)
(2011)
\end{botherref}
\endbibitem

\bibitem[\protect\citeauthoryear{Wang et~al.}{2018}]{wang2018graphgan}
\begin{bchapter}
\bauthor{\bsnm{Wang}, \binits{H.}},
\bauthor{\bsnm{Wang}, \binits{J.}},
\bauthor{\bsnm{Wang}, \binits{J.}},
\bauthor{\bsnm{Zhao}, \binits{M.}},
\bauthor{\bsnm{Zhang}, \binits{W.}},
\bauthor{\bsnm{Zhang}, \binits{F.}},
\bauthor{\bsnm{Xie}, \binits{X.}},
\bauthor{\bsnm{Guo}, \binits{M.}}:
\bctitle{Graphgan: Graph representation learning with generative adversarial nets}.
In: \bbtitle{Proceedings of the AAAI Conference on Artificial Intelligence},
vol. \bseriesno{32}
(\byear{2018})
\end{bchapter}
\endbibitem

\bibitem[\protect\citeauthoryear{Qiu et~al.}{2018}]{qiu2018network}
\begin{bchapter}
\bauthor{\bsnm{Qiu}, \binits{J.}},
\bauthor{\bsnm{Dong}, \binits{Y.}},
\bauthor{\bsnm{Ma}, \binits{H.}},
\bauthor{\bsnm{Li}, \binits{J.}},
\bauthor{\bsnm{Wang}, \binits{K.}},
\bauthor{\bsnm{Tang}, \binits{J.}}:
\bctitle{Network embedding as matrix factorization: Unifying deepwalk, line, pte, and node2vec}.
In: \bbtitle{Proceedings of the Eleventh ACM International Conference on Web Search and Data Mining},
pp. \bfpage{459}--\blpage{467}
(\byear{2018})
\end{bchapter}
\endbibitem

\bibitem[\protect\citeauthoryear{Hamilton et~al.}{2017}]{hamilton2017inductive}
\begin{botherref}
\oauthor{\bsnm{Hamilton}, \binits{W.}},
\oauthor{\bsnm{Ying}, \binits{Z.}},
\oauthor{\bsnm{Leskovec}, \binits{J.}}:
Inductive representation learning on large graphs.
Advances in neural information processing systems
\textbf{30}
(2017)
\end{botherref}
\endbibitem

\bibitem[\protect\citeauthoryear{Kipf and Welling}{2016}]{kipf2016variational}
\begin{botherref}
\oauthor{\bsnm{Kipf}, \binits{T.N.}},
\oauthor{\bsnm{Welling}, \binits{M.}}:
Variational graph auto-encoders.
arXiv preprint arXiv:1611.07308
(2016)
\end{botherref}
\endbibitem

\bibitem[\protect\citeauthoryear{You et~al.}{2018}]{you2018graphrnn}
\begin{bchapter}
\bauthor{\bsnm{You}, \binits{J.}},
\bauthor{\bsnm{Ying}, \binits{R.}},
\bauthor{\bsnm{Ren}, \binits{X.}},
\bauthor{\bsnm{Hamilton}, \binits{W.}},
\bauthor{\bsnm{Leskovec}, \binits{J.}}:
\bctitle{Graphrnn: Generating realistic graphs with deep auto-regressive models}.
In: \bbtitle{International Conference on Machine Learning},
pp. \bfpage{5708}--\blpage{5717}
(\byear{2018}).
\bcomment{PMLR}
\end{bchapter}
\endbibitem

\bibitem[\protect\citeauthoryear{She and Chai}{2017}]{she2017interactive}
\begin{bchapter}
\bauthor{\bsnm{She}, \binits{L.}},
\bauthor{\bsnm{Chai}, \binits{J.}}:
\bctitle{Interactive learning of grounded verb semantics towards human-robot communication}.
In: \bbtitle{Proceedings of the 55th Annual Meeting of the Association for Computational Linguistics (Volume 1: Long Papers)},
pp. \bfpage{1634}--\blpage{1644}
(\byear{2017})
\end{bchapter}
\endbibitem

\bibitem[\protect\citeauthoryear{Jin et~al.}{2020}]{jin2020graph}
\begin{bchapter}
\bauthor{\bsnm{Jin}, \binits{W.}},
\bauthor{\bsnm{Ma}, \binits{Y.}},
\bauthor{\bsnm{Liu}, \binits{X.}},
\bauthor{\bsnm{Tang}, \binits{X.}},
\bauthor{\bsnm{Wang}, \binits{S.}},
\bauthor{\bsnm{Tang}, \binits{J.}}:
\bctitle{Graph structure learning for robust graph neural networks}.
In: \bbtitle{Proceedings of the 26th ACM SIGKDD International Conference on Knowledge Discovery \& Data Mining},
pp. \bfpage{66}--\blpage{74}
(\byear{2020})
\end{bchapter}
\endbibitem

\bibitem[\protect\citeauthoryear{Cherian and Sullivan}{2019}]{cherian2019sem}
\begin{bchapter}
\bauthor{\bsnm{Cherian}, \binits{A.}},
\bauthor{\bsnm{Sullivan}, \binits{A.}}:
\bctitle{Sem-gan: Semantically-consistent image-to-image translation}.
In: \bbtitle{2019 Ieee Winter Conference on Applications of Computer Vision (wacv)},
pp. \bfpage{1797}--\blpage{1806}
(\byear{2019}).
\bcomment{IEEE}
\end{bchapter}
\endbibitem

\bibitem[\protect\citeauthoryear{Fraccaro et~al.}{2016}]{fraccaro2016sequential}
\begin{botherref}
\oauthor{\bsnm{Fraccaro}, \binits{M.}},
\oauthor{\bsnm{S{\o}nderby}, \binits{S.K.}},
\oauthor{\bsnm{Paquet}, \binits{U.}},
\oauthor{\bsnm{Winther}, \binits{O.}}:
Sequential neural models with stochastic layers.
Advances in neural information processing systems
\textbf{29}
(2016)
\end{botherref}
\endbibitem

\bibitem[\protect\citeauthoryear{Brin and Page}{1998}]{brin1998anatomy}
\begin{barticle}
\bauthor{\bsnm{Brin}, \binits{S.}},
\bauthor{\bsnm{Page}, \binits{L.}}:
\batitle{The anatomy of a large-scale hypertextual web search engine}.
\bjtitle{Computer networks and ISDN systems}
\bvolume{30}(\bissue{1-7}),
\bfpage{107}--\blpage{117}
(\byear{1998})
\end{barticle}
\endbibitem

\bibitem[\protect\citeauthoryear{Fortunato}{2010}]{fortunato2010community}
\begin{barticle}
\bauthor{\bsnm{Fortunato}, \binits{S.}}:
\batitle{Community detection in graphs}.
\bjtitle{Physics reports}
\bvolume{486}(\bissue{3-5}),
\bfpage{75}--\blpage{174}
(\byear{2010})
\end{barticle}
\endbibitem

\bibitem[\protect\citeauthoryear{Yang and Leskovec}{2014}]{yang2014overlapping}
\begin{barticle}
\bauthor{\bsnm{Yang}, \binits{J.}},
\bauthor{\bsnm{Leskovec}, \binits{J.}}:
\batitle{Overlapping communities explain core--periphery organization of networks}.
\bjtitle{Proceedings of the IEEE}
\bvolume{102}(\bissue{12}),
\bfpage{1892}--\blpage{1902}
(\byear{2014})
\end{barticle}
\endbibitem

\bibitem[\protect\citeauthoryear{Henderson et~al.}{2012}]{henderson2012rolx}
\begin{bchapter}
\bauthor{\bsnm{Henderson}, \binits{K.}},
\bauthor{\bsnm{Gallagher}, \binits{B.}},
\bauthor{\bsnm{Eliassi-Rad}, \binits{T.}},
\bauthor{\bsnm{Tong}, \binits{H.}},
\bauthor{\bsnm{Basu}, \binits{S.}},
\bauthor{\bsnm{Akoglu}, \binits{L.}},
\bauthor{\bsnm{Koutra}, \binits{D.}},
\bauthor{\bsnm{Faloutsos}, \binits{C.}},
\bauthor{\bsnm{Li}, \binits{L.}}:
\bctitle{Rolx: structural role extraction \& mining in large graphs}.
In: \bbtitle{Proceedings of the 18th ACM SIGKDD International Conference on Knowledge Discovery and Data Mining},
pp. \bfpage{1231}--\blpage{1239}
(\byear{2012})
\end{bchapter}
\endbibitem

\bibitem[\protect\citeauthoryear{Le and Mikolov}{2014}]{le2014distributed}
\begin{bchapter}
\bauthor{\bsnm{Le}, \binits{Q.}},
\bauthor{\bsnm{Mikolov}, \binits{T.}}:
\bctitle{Distributed representations of sentences and documents}.
In: \bbtitle{International Conference on Machine Learning},
pp. \bfpage{1188}--\blpage{1196}
(\byear{2014}).
\bcomment{PMLR}
\end{bchapter}
\endbibitem

\bibitem[\protect\citeauthoryear{Sun et~al.}{2011}]{sun2011pathsim}
\begin{barticle}
\bauthor{\bsnm{Sun}, \binits{Y.}},
\bauthor{\bsnm{Han}, \binits{J.}},
\bauthor{\bsnm{Yan}, \binits{X.}},
\bauthor{\bsnm{Yu}, \binits{P.S.}},
\bauthor{\bsnm{Wu}, \binits{T.}}:
\batitle{Pathsim: Meta path-based top-k similarity search in heterogeneous information networks}.
\bjtitle{Proceedings of the VLDB Endowment}
\bvolume{4}(\bissue{11}),
\bfpage{992}--\blpage{1003}
(\byear{2011})
\end{barticle}
\endbibitem

\bibitem[\protect\citeauthoryear{Rabin and Scott}{1959}]{5392601}
\begin{barticle}
\bauthor{\bsnm{Rabin}, \binits{M.O.}},
\bauthor{\bsnm{Scott}, \binits{D.}}:
\batitle{Finite automata and their decision problems}.
\bjtitle{IBM Journal of Research and Development}
\bvolume{3}(\bissue{2}),
\bfpage{114}--\blpage{125}
(\byear{1959})
\doiurl{10.1147/rd.32.0114}
\end{barticle}
\endbibitem

\bibitem[\protect\citeauthoryear{Friedl}{2006}]{friedl2006mastering}
\begin{bbook}
\bauthor{\bsnm{Friedl}, \binits{J.E.}}:
\bbtitle{Mastering Regular Expressions}.
\bpublisher{" O'Reilly Media, Inc."}, \blocation{???}
(\byear{2006})
\end{bbook}
\endbibitem

\bibitem[\protect\citeauthoryear{Lee and Yannakakis}{1996}]{lee1996principles}
\begin{barticle}
\bauthor{\bsnm{Lee}, \binits{D.}},
\bauthor{\bsnm{Yannakakis}, \binits{M.}}:
\batitle{Principles and methods of testing finite state machines-a survey}.
\bjtitle{Proceedings of the IEEE}
\bvolume{84}(\bissue{8}),
\bfpage{1090}--\blpage{1123}
(\byear{1996})
\end{barticle}
\endbibitem

\end{thebibliography}

\end{document}